\documentclass[10pt,letterpaper]{article}
\usepackage[T1]{fontenc}
\usepackage[letterpaper, margin=1in, top=1.2in, bottom=1.2in]{geometry}
\usepackage{times}
\usepackage{helvet}
\usepackage{microtype}

\usepackage{amsmath}
\usepackage{amsfonts}
\usepackage{amssymb}
\usepackage{graphicx}
\usepackage[numbers,sort&compress]{natbib}
\usepackage{enumitem}
\usepackage{pifont} 
\usepackage{fontawesome5}
\usepackage{nicefrac}
\usepackage[table]{xcolor}

\usepackage{booktabs}
\usepackage{caption}
\usepackage{subcaption}
\usepackage{multirow}
\usepackage{makecell}
\usepackage{arydshln}
\usepackage{longtable}
\usepackage{tabularx}
\usepackage{array}
\usepackage{colortbl}
\usepackage{wrapfig}
\usepackage{placeins}

\usepackage{algorithm}
\usepackage{algpseudocode}
\usepackage{fvextra} 

\usepackage{tcolorbox}
\tcbuselibrary{skins, breakable}
\usepackage[framemethod=tikz]{mdframed}
\usepackage{titlesec}
\usepackage{needspace} 
\usepackage{authblk}
\usepackage{fancyhdr}
\usepackage{tikz}

\usepackage{hyperref}
\usepackage{cleveref}
\Crefname{figure}{Figure}{Figures}
\Crefname{table}{Table}{Tables}
\Crefname{section}{Section}{Sections}
\Crefname{equation}{Equation}{Equations}
\Crefname{algorithm}{Algorithm}{Algorithms}
\Crefname{appendix}{Appendix}{Appendix}

\definecolor{darkblue}{RGB}{0, 51, 102}
\definecolor{lightbluebg}{RGB}{245, 248, 252}
\definecolor{blueframe}{RGB}{90, 150, 200}
\definecolor{dividercolor}{RGB}{200, 210, 220}
\definecolor{yellowtext}{RGB}{68,132,243}
\definecolor{yellowred}{RGB}{50,167,82}
\definecolor{yellowblue}{RGB}{251,191,5}
\definecolor{darkgreen}{rgb}{0,0.4,0}
\definecolor{maroon}{HTML}{A00000}
\definecolor{gray}{rgb}{0.5, 0.5, 0.5}
\definecolor{chocolate}{HTML}{D2691E}
\definecolor{indigo}{HTML}{4B0082}
\definecolor{violet}{HTML}{4B2E83}
\definecolor{lightgreen}{HTML}{E0FBE0}
\definecolor{lightred}{HTML}{FBE0E0}
\definecolor{cadmiumgreen}{rgb}{0.0, 0.42, 0.24}
\definecolor{forestgreen}{rgb}{0.13, 0.55, 0.13}
\definecolor{lightgray}{rgb}{0.9, 0.9, 0.9}
\definecolor{exp_table_blue}{HTML}{DAECED}
\definecolor{exp_table_yelow}{HTML}{F0B100}  
\definecolor{exp_table_red}{HTML}{CF292B} 
\definecolor{exp_table_green}{HTML}{05B04F} 

\hypersetup{
    colorlinks=true,
    linkcolor=darkblue,
    citecolor=darkgreen,
    urlcolor=darkblue
}

\titleformat{\section}{\sffamily\Large\bfseries\color{darkblue}}{\thesection}{1em}{}
\titleformat{\subsection}{\sffamily\large\bfseries\color{darkblue}}{\thesubsection}{1em}{}
\titleformat{\subsubsection}{\sffamily\normalsize\bfseries\color{darkblue}}{\thesubsubsection}{1em}{}
\titleformat{\paragraph}[runin]{\sffamily\normalsize\bfseries}{}{0em}{}[]
\titlespacing{\paragraph}{0pt}{0.5ex plus 0.2ex minus 0.1ex}{0.5em}

\setlist[enumerate,itemize]{topsep=0pt, itemsep=0pt, leftmargin=*, after=\leavevmode}
\setlist[enumerate,1]{label=(\arabic*)}

\DefineVerbatimEnvironment{VerbatimWrap}{Verbatim}{
    breaklines=true,
    breakindent=0pt,
    breaksymbol={},
    fontsize=\scriptsize,
}

\mdfdefinestyle{customstyle}{
    linecolor=cyan!70,
    linewidth=3pt,
    innerleftmargin=5pt,
    topline=false,
    rightline=false,
    bottomline=false,
    leftline=true,
    innerrightmargin=5pt,
    innertopmargin=5pt,
    innerbottommargin=5pt,
    backgroundcolor=cyan!8,
}

\newtcolorbox{prompt}[1]{
    colback=lightbluebg!30!white,
    colframe=blueframe,
    breakable,
    title=\textit{#1}
}

\makeatletter
\def\@title{}
\def\@abstract{}
\def\@keywords{}
\def\@codelink{}
\def\@datasetlink{}
\def\@projectlink{}
\def\@footnotemarkers{}

\renewcommand{\title}[1]{\def\@title{#1}}
\renewcommand{\author}[1]{\def\@author{#1}}
\renewcommand{\abstract}[1]{\def\@abstract{#1}}
\newcommand{\keywords}[1]{\def\@keywords{#1}}
\newcommand{\codelink}[1]{\def\@codelink{#1}}
\newcommand{\datasetlink}[1]{\def\@datasetlink{#1}}
\newcommand{\projectlink}[1]{\def\@projectlink{#1}}
\newcommand{\footnotemarkers}[1]{\def\@footnotemarkers{#1}}

\fancypagestyle{firstpage}{
  \fancyhf{}
  \ifx\@footnotemarkers\@empty
    \fancyfoot[C]{\thepage}
  \else
    \fancyfoot[L]{\rule{0.333\textwidth}{0.4pt}\\[2pt]\footnotesize \@footnotemarkers}
  \fi
  
}

\renewcommand{\maketitle}{%
  \thispagestyle{firstpage}%
  \begin{tcolorbox}[
    colback=lightbluebg,
    colframe=blueframe,
    arc=3mm,
    boxrule=0.5pt,
    left=17pt,
    right=17pt,
    top=17pt,
    bottom=17pt,
    width=\textwidth,
    boxsep=3pt
  ]
    \noindent
    {\sffamily\fontsize{18}{22}\selectfont\bfseries\color{darkblue}\raggedright\@title\par}
    \vspace{12pt}
    \noindent
    {\sffamily\fontsize{9.5}{12}\selectfont\raggedright\@author\par}
    \vspace{7pt}
    {\color{dividercolor}\rule{\linewidth}{0.5pt}}\par
    \vspace{7pt}
    \ifx\@abstract\@empty\else
      \noindent\parbox{\linewidth}{%
        {\sffamily\bfseries Abstract}\par\smallskip
        {\fontsize{10.5}{13.2}\selectfont\@abstract\par}%
      }\par
      \vspace{1ex} 
    \fi
    \ifx\@keywords\@empty\else
      \noindent\parbox{\linewidth}{%
        {\sffamily\bfseries Keywords:}\quad \@keywords%
      }\par
      \vspace{1.5ex} 
    \fi
    \ifx\@codelink\@empty
      \ifx\@datasetlink\@empty
        \ifx\@projectlink\@empty
        \else
          \vspace{-1.1em}
          {\color{dividercolor}\rule{\linewidth}{0.5pt}}
          \vspace{-0.6em}
          \centering
          \sffamily\small
          \href{\@projectlink}{\faGlobe\ Project Page}
          \par
        \fi
      \else
        \vspace{-1.1em}
        {\color{dividercolor}\rule{\linewidth}{0.5pt}}
        \vspace{-0.6em}
        \centering
        \sffamily\small
        \ifx\@projectlink\@empty\else
          \href{\@projectlink}{\faGlobe\ Project Page}\hspace{1.5em}
        \fi
        \href{\@datasetlink}{\faDatabase\ Dataset}
        \par
      \fi
    \else
      \vspace{-1.1em}
      {\color{dividercolor}\rule{\linewidth}{0.5pt}}
      \vspace{-0.6em}
      \centering
      \sffamily\small
      \href{\@codelink}{\faGithub\ Code}%
      \ifx\@datasetlink\@empty\else
        \hspace{1.5em}\href{\@datasetlink}{\faDatabase\ Dataset}%
      \fi
      \ifx\@projectlink\@empty\else
        \hspace{1.5em}\href{\@projectlink}{\faGlobe\ Project Page}%
      \fi
      \par
    \fi
  \end{tcolorbox}
}
\makeatother
\usetikzlibrary{arrows.meta,positioning,calc}
\newcommand{\leadpara}[1]{\noindent\textbf{#1}\ }
\newcommand{\sysname}{CarryOn}


\hypersetup{
  pdftitle={CarryOn: A Dual-Loop Harness for Personalized and Reliable Robot Assistance},
  pdfauthor={Zixuan Wang, Yuqi Liu, Yuxin Chen, Sengqiao Yang, Jinhui Ye, Pengguang Chen, Changsheng Lu, Shu Liu, Bei Yu, Jiaya Jia},
  pdfsubject={Personalized robot assistance}
}

\usepackage[normalem]{ulem}
\newif\ifmethodreview
\methodreviewfalse
\definecolor{methodadded}{RGB}{195,25,25}
\definecolor{methodremoved}{RGB}{30,85,165}

\newcommand{\methoddel}[1]{\ifmethodreview{\color{methodremoved}\sout{#1}}\fi}

\title{\sysname{}: A Dual-Loop Harness for Personalized and Reliable Robot Assistance}
\author{%
    \mbox{\textbf{Zixuan Wang}$^{1*}$}\hspace{10pt}%
    \mbox{\textbf{Yuqi Liu}$^{2*}$}\hspace{10pt}%
    \mbox{\textbf{Yuxin Chen}$^{1*}$}\hspace{10pt}%
    \mbox{\textbf{Sengqiao Yang}$^{2}$}\hspace{10pt}%
    \mbox{\textbf{Jinhui Ye}$^{1}$}\\[3pt]
    \mbox{\textbf{Pengguang Chen}$^{3}$}\hspace{10pt}%
    \mbox{\textbf{Changsheng Lu}$^{1}$}\hspace{10pt}%
    \mbox{\textbf{Shu Liu}$^{3}$}\hspace{10pt}%
    \mbox{\textbf{Bei Yu}$^{2}$}\hspace{10pt}%
    \mbox{\textbf{Jiaya Jia}$^{1,3}$}\\[6pt]
    HKUST$^{1}$\hspace{18pt}CUHK$^{2}$\hspace{18pt}SmartMore$^{3}$\\[3pt]
    {\ttfamily\small\url{https://github.com/JIA-Lab-research/CarryOn}}
}
\date{}
\abstract{
While generalist vision-language-action (VLA) models have enabled diverse open-world manipulation, real-world robotic assistance requires two critical capabilities they currently lack: personalization and reliable execution. First, executing an instruction like \textit{`make me a sandwich'} requires grounding subtask decomposition in persistent user preferences rather than relying on a single generic solution. Second, reliable deployment demands explainable safety assurance that proactively prevents hazards and gracefully recovers from occasional physical failures.
To address these gaps, we present \sysname{}, a dual-loop harness designed to augment a frozen VLA policy. The planning loop maintains persistent user-level and scene-level context that is continually grounded against new observations and revised through interaction, enabling personalized task selection and execution grounding across conversations and changing scenes. The execution loop actively supervises policy rollouts, shielding actions from predicted hazardous contact. When execution fails, the system invokes reusable control-based recovery skills that restore task progress and are refined through reflection on earlier failures. On a real robot and in simulation, \sysname{} dispatches preference-consistent subtasks across users and scenes, reduces the violation rate of an already safety-finetuned VLA from 52.6\% to 6.1\%, and increases recovery success from 13\% to 80\% when refined skills are reused.
}

\begin{document}
\newgeometry{left=0.9in,right=0.9in,top=0.65in,bottom=0.65in}
\maketitle
\begin{figure}[H]
    \centering
    \includegraphics[width=0.78\linewidth]{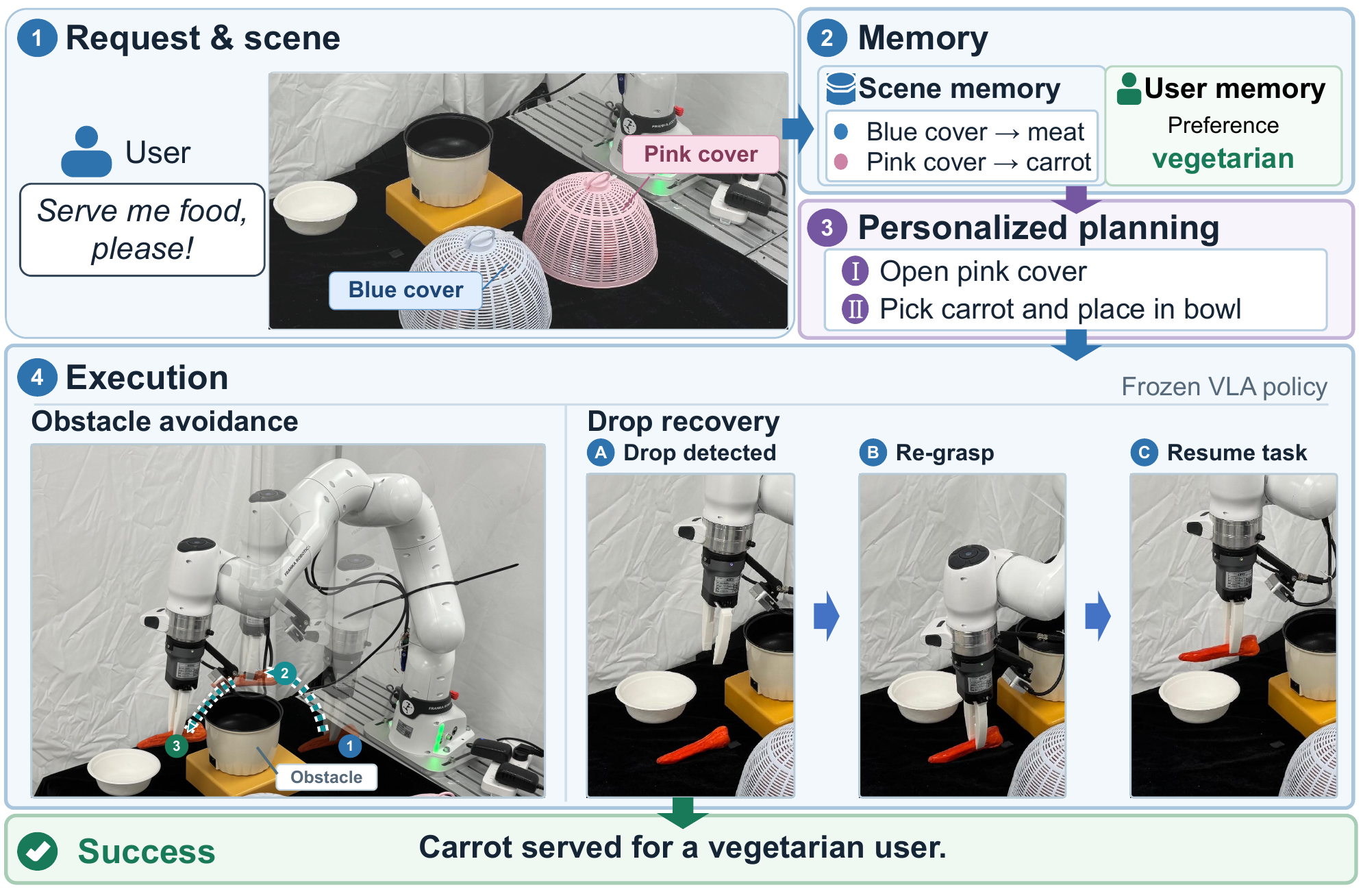}
    \caption{Serving food to a vegetarian user. (1) The user asks only, ``Serve me food.'' (2) \sysname{} retrieves the context from memory. (3) The planner dispatches a sequence of subtasks to a frozen VLA policy: open the pink cover and, if the carrot is underneath, place it in the bowl. (4) During execution, a proactive shield steers the gripper and the carried carrot around an obstacle, and reactive recovery re-grasps the carrot to resume the task.}
    \label{fig:teaser}
\end{figure}

\newgeometry{left=1in,right=1in,top=1in,bottom=1in}
\section{Introduction}
\label{sec:introduction}
Robot assistance ultimately aims to serve people in their everyday lives. Generalist vision-language-action (VLA) models have advanced this goal by learning to execute diverse manipulation instructions across tasks and environments~\cite{black2024pi_0,kim2024openvla,brohan2023rt}. Large-scale robot datasets further support learning across tasks and embodiments~\cite{o2024open,team2024octo,bousmalis2023robocat}. However, progress in instruction-conditioned manipulation alone does not bridge the gap to everyday assistance. A robot must know both \emph{what to do} to meet the user's needs in the current environment and \emph{how to carry it out reliably} as physical conditions change. Bridging this gap requires a system that connects context-sensitive task decisions with reliable execution.

A \emph{harness} is the system surrounding a base model that orchestrates execution and decides how the model plans, perceives and manages context, stores artifacts, and evaluates results~\cite{weng2026harness}. Language-agent systems illustrate how memory and reflection can support decisions across interactions~\cite{packer2023memgpt,park2023generative}. In the field of robotics, hierarchical VLA systems connect high-level requests to manipulation instructions~\cite{shi2025hi}. Recent concurrent works on embodied harnesses use memory to steer frozen VLAs and coordinate heterogeneous policies~\cite{zhang2026harness,huang2026roboharness}, or evolve runtime critics and recovery skills from execution experience~\cite{ding2026zetta}. These works mainly focus on how to improve VLA models' performance in out-of-distribution (OOD) scenarios. However, for everyday assistance, reliable execution must also be coupled with deciding what is appropriate for the user in the current situation.

We study this connection through personalized robot assistance, which exposes two challenges. First, task decisions must remain aligned with the user's needs as interactions and scenes change, and the same request can require different actions. Learning preferences or identifying personal objects~\cite{wu2023tidybot,lee2025bring} provides context, while in reality the context can be dynamically changed and thus requires revision in stored memory. Second, the resulting plan must be supported by reliable physical execution: impending hazardous contact calls for intervention~\cite{hu2025vlsa} to ensure safety, while a failure that has already occurred calls for repair~\cite{vats2025recoverychaining} to ensure robustness. The harness must connect these responses to the intended task and retain failure experience to improve later recovery.

Consider a user who asks a robot to ``serve me food'' (Fig.~\ref{fig:teaser}). The robot should respect an earlier vegetarian preference without requiring it to be repeated in the query. It may look beneath a remembered cover for a carrot, but finding meat instead must change its original plan. Even when the ingredient is suitable, an obstacle or a lost grasp can interrupt delivery to the bowl. \emph{The harness must preserve the user's dietary requirement when observations change the plan, restore progress when a grasp fails, and use that failure to improve later recovery.}

We present \sysname{}, a \emph{dual-loop harness} that connects adaptive task decisions with reliable execution for personalized robot assistance. Its central design makes task context and recovery behavior persistent, revisable parts of the runtime around a frozen VLA. The planning loop determines what to do for the user and scene, and the execution loop protects and repairs the active subtask. Visual feedback connects them, informing plan revision and determining whether execution should advance or invoke recovery while retaining the task goal.

The \emph{planning loop} supports contextual adaptation through memories. User context expresses needs and preferences, while scene context helps determine where and how they can be fulfilled. The planner combines both with the current multimodal request and available observations to select an endpoint and supported subtasks. Dialogue revises user context, and observations trigger explicit scene-memory updates before subsequent plans proceed. Memory thus participates in a feedback loop: it guides action, while the resulting evidence updates what later plans can rely on.

The \emph{execution loop} supports physical reliability through fast proactive shielding and slower reactive recovery. A fast proactive shield predicts gripper and carried-object motion and filters commands against geometric contact constraints. After a failure, slower, event-triggered recovery skills repair the physical state so the interrupted task can continue. Offline reflection then uses failures of the current recovery version to revise its parameters and strategies, retaining successive versions in a skill library for later attempts. Immediate protection and reactive repair operate within an episode, and accumulated experience changes recovery behavior across attempts. This design keeps the VLA fixed while allowing the recovery pathway to improve through repeated reflection.

\textbf{Contributions.} In this paper, we present a dual-loop harness, called \sysname{}, with the features of personalization and reliable execution. The \textit{planning loop} of the harness connects persistent user and scene contexts with observation-driven revision, supporting personalized task decisions across interactions, scenes, and query modalities, while the \textit{execution loop} combines proactive geometric shielding with reusable and iteratively refined recovery skills. Real-world studies demonstrate the effectiveness of our harness in planning preference-oriented tasks across diverse users and scenes. Notably, our method strongly increases recovery success rates from 13\% to 80\% using refined skills, and reduces the constraint-violation rate from 52.6\% to 6.1\%.


\section{Related Work}
\label{sec:related_work}

\leadpara{Personalization through persistent context.}
TidyBot learns placement preferences that generalize to later requests~\cite{wu2023tidybot}, while Bring My Cup personalizes through reference-image memory and visual prompting~\cite{lee2025bring}. Embodied personalization studies memory across user interactions~\cite{kwon2026embodied,lee2026personalizing}, and language agents retrieve and consolidate experience for subsequent decisions~\cite{packer2023memgpt,park2023generative}. Unlike previous works, \sysname{} connects persistent preferences with revisable scene knowledge: user memory constrains task selection, while the scene memory guides later actions.

\leadpara{Harnesses for robot policy execution.}
Code as Policies generates programs over perception and control APIs~\cite{liang2023code}. Language-model planners ground instructions through affordances, feedback, and multimodal observations~\cite{ahn2022can,singh2022progprompt,huang2022inner,huang2023voxposer,driess2023palm,zhu2023vima}, while uncertainty-aware planners decide when to act or seek information~\cite{ren2023robots}. Hierarchical VLA systems connect instructions to visuomotor execution~\cite{shi2025hi}; RoboHarness uses memory to orchestrate heterogeneous policies~\cite{huang2026roboharness}. Harness VLA combines a frozen contact-rich primitive with analytic operations and memory-guided retries~\cite{zhang2026harness}. In contrast, \sysname{} retains the VLA to execute manipulation and couples persistent personalization with action filtering and physical repair. Both loops share the active task, preserving the user-conditioned goal through recovery and using observed outcomes to govern task progression.

\leadpara{Recovery and adaptation from failure.}
Vision-language models provide semantic failure feedback~\cite{zhang2026safevla}, while FailSafe trains failure reasoning and recovery from generated failure recovery cases~\cite{lin2025failsafe}. RecoveryChaining learns local policies that return control to nominal controllers~\cite{vats2025recoverychaining}, and Zetta evolves runtime critics and recovery skills around a frozen policy~\cite{ding2026zetta}. Language agents accumulate experience through memory and reusable skills~\cite{shinn2023reflexion,zhao2024expel,wang2023voyager}. In our method, \sysname{} confines refinement to control-based recovery: offline reflection revises supported parameters and strategies, retaining successive skills for later attempts. This separates immediate repair from experience accumulation while leaving the VLA and geometric shield fixed.

\leadpara{Geometric protection during manipulation.}
Latent-space safety filters use reachability analysis in learned world models~\cite{nakamura2025generalizing}; AnySafe adapts such filters to parameterized constraints~\cite{agrawal2025anysafe}. VLSA adds a barrier-function safety layer~\cite{hu2025vlsa}, and Code-as-Monitor generates executable constraints for failure monitoring~\cite{zhou2025code}. In our design, the proactive shield predicts the spatial response to policy-predicted actions and filters actions according to minimum-distance constraints between the gripper or carried object and task-specific obstacles. This provides a check before hazardous contact, complementing recovery after failure without online learning.

\section{Method}
\label{sec:method}

\begin{figure}[htbp]
    \centering
    \includegraphics[width=1.\linewidth]{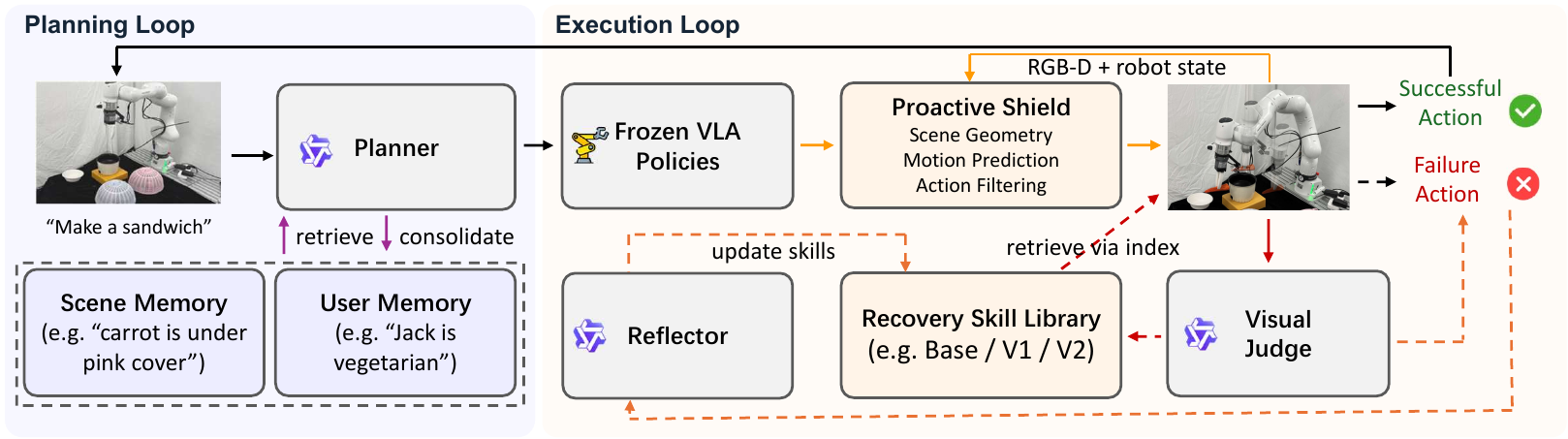}
    \caption{\textbf{Two loops, one task context.} The planning loop grounds a request in persistent context and dispatches supported subtasks. The execution loop retains the VLA for manipulation, with proactive shielding before hazardous contact and recovery after failure. A shared visual judge advances the plan on success or triggers local recovery on failure.  Recovery attempts to repair the physical state so the VLA can continue the interrupted subtask. Post-episode reflection uses failure evidence to revise recovery skills, retaining successive versions in the skill library.}
    \label{fig:architecture}
\end{figure}

\sysname{} uses a dual-loop harness to adapt task decisions to the user and environment and to support reliable physical execution. The planning loop turns a request into subtasks using persistent user and scene knowledge, then updates that knowledge through dialogue and observation. The execution loop runs a pretrained VLA, checks its actions for collision risk, and invokes control-based recovery skills when manipulation fails. Experience is retained in two forms: context for future planning and recovery experience that supports successive revisions of a versioned skill library. Fig.~\ref{fig:architecture} summarizes the runtime, and the following sections describe how each form of experience is used and revised.

\subsection{Overview of \sysname{}}

Given a request $c$ and an optional reference image, the planner combines user memory $M^u$, scene memory $M^s$, and descriptions of the available skills to produce a plan. Each planned subtask specifies an instruction $\ell_i$, an endpoint that provides the requested skill, and optional dependencies on earlier subtasks. Endpoints include manipulation stations and information-gathering tools, so a plan can request an observation or external information before deciding what to manipulate. At a manipulation endpoint, the frozen VLA predicts action chunks $U_t=\pi_\theta(o_t,\ell_i)$ from observation $o_t$ and the subtask instruction $\ell_i$. A geometric shield filters these chunks using task-dependent contact constraints before execution. The Visual Judge compares recent camera images with the scene at the start of the subtask to assess progress toward $\ell_i$. It reports \emph{running}, \emph{success}, or \emph{failed}: success advances the plan, while failure invokes recovery on the same subtask. Observation results return to the planner for plan and memory updates.

\Needspace{6\baselineskip}
\subsection{The Planning Loop: Personalization across Interactions and Scenes}
\label{sec:method-know}

\subsubsection{Planning with persistent, revisable context}

A personalized plan must satisfy the user's needs and account for the environment in which the task will be carried out. Since the current request may leave both kinds of information unstated, we maintain user and scene memories as complementary planning context. User memory supplies preferences and object references established in earlier conversations. We store preferences, habits, and reference images of the user's objects in user-specific profiles, which enter the planner context with each request. The planner is instructed to apply the requesting user's profile when selecting subtasks. The profile is combined with the current instruction, so planning accounts for both retained preferences and the user's present request.

Scene memory provides planning context for deciding where and how to carry out the task in the current environment. We store object inventories and descriptions by place. The planner combines these records with the available skill descriptions to select an endpoint and decide whether an observation is needed before manipulation. If the target is hidden or its location needs checking, the planner requests an observation. The returned facts enter the planning context and are combined with the user context to revise subsequent subtasks.

\subsubsection{Updating context from dialogue and observation}

Personalized assistance must adapt as user needs and the surrounding environment change. This requires the robot to distinguish lasting changes in user preferences from temporary requests, and to revise its understanding of the scene as new evidence becomes available. We update user memory from dialogue and scene memory from observations to keep this planning context current. For user memory, the planner is instructed to store lasting preferences during the interaction and keep temporary instructions in the current conversation. After a plan terminates, a background summarizer reviews the dialogue using the same criterion to retain user facts not yet recorded. When the user revises or retracts a preference, the corresponding profile entry is replaced or removed, so later plans use the updated requirements.

The environment can also change independently of the user's instructions. New observations must therefore revise the robot's understanding of the scene, both for the current task and for later requests. We require a scene-memory update whenever an observation reveals the target or finds its remembered location empty or stale. The runtime retains the observation and records the need for an update. The planner uses this evidence to specify an \texttt{update\_scene} operation in its next plan, revising the scene record and removing outdated information. If the planner omits this required operation, a plan validator returns the plan to the planner for revision.

\subsection{The Execution Loop: Proactive Protection and Reactive Recovery}
\label{sec:method-act}

\begin{figure}[!htbp]
\centering
\includegraphics[width=0.8\linewidth]{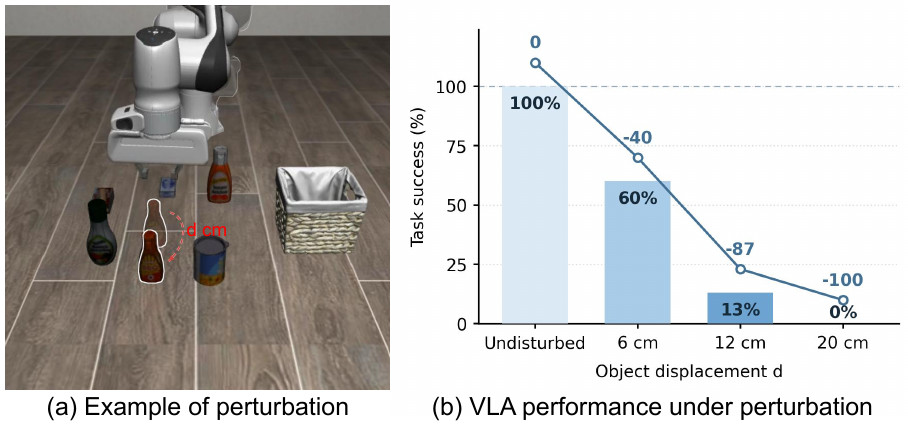}
\caption{(a) Schematic of the perturbation, varying object displacement relative to the initial lifted position. 
(b) Policy-only response to object displacement on LIBERO-Object. Blue bars show task success, and the upper line shows the decrease from undisturbed success in percentage points.}
\label{fig:disturbance-response}
\end{figure}

 Reliable physical assistance requires both \emph{safety} and \emph{robustness}: the robot must intervene before hazardous contact and restore progress after disturbances interrupt a task. A policy trained to complete the task may still fall short of either requirement. Two observations motivate intervention around the VLA: The safety-finetuned $\pi_{0.5}$ checkpoint, trained on 19,664 safe demonstrations~\cite{cui2026libero}, still violates constraints in 52.6\% of our 900 evaluation episodes, compared with 71.1\% for the original checkpoint (Section~\ref{sec:safety}). In a separate LIBERO-Object policy-only probe, frozen StarVLA~\cite{community2026starvla} completes 26/200 episodes (13.0\%) after a grasped object is displaced by 12\,cm (Fig.~\ref{fig:disturbance-response}). 

We therefore divide execution support into fast \emph{proactive shielding} and slower, event-triggered \emph{reactive recovery}. The shield checks actions predicted by the VLA before execution and filters them according to predicted geometric contact risk. Shielding reduces risks in the proposed motion, but surrounding environment may still introduce disturbance and change the physical state. Reactive recovery acts after the Judge detects a subtask failure, suspending the VLA and invoking a control-based skill to attempt repair before continue the same subtask. Both interventions operate within an episode, at action and subtask levels respectively. Offline reflection subsequently uses unresolved failures to refine the current recovery version for later attempts, while the VLA remains fixed.

\subsubsection{Proactive shielding within an episode}

A VLA policy may predict actions that cause unintended contact even when the task instruction itself is appropriate. Safe execution therefore requires checking the physical consequences of these actions before execution. We implement proactive protection with a geometric action filter, the \emph{shield}, between the policy and controller. For each action chunk, it uses current RGB-D observations and the target and destination descriptions to construct obstacle sets, predict gripper and carried-object motion, and filter the actions. The output is an executable action prefix or, if the first action is blocked, a short retreat.
\label{sec:method-proactive}

Grasping and placement require contact with the target and destination, so treating every scene point as an obstacle would prevent the task itself. We construct separate obstacle sets for the gripper and carried object by identifying which observed geometry belongs to these task roles. A VLM infers descriptions of the target and destination from the task instruction. The scene builder then uses text-prompted segmentation to associate these descriptions with image masks and projects the masked depth pixels into three-dimensional point sets. It removes robot self-geometry using the robot model and current state and identifies the target's local supporting surface from geometry beneath it. Contact rules exclude permitted target, support, and destination geometry from each body's obstacle set. Other points, including unassigned geometry, remain obstacles. Appendix~\ref{app:shield-grounding} describes the grounding and contact rules used in the safety evaluation.

Motion prediction then estimates the end-effector trajectory resulting from the VLA's predicted actions. Proposed action increments alone do not describe the realized robot displacement: the controller response and ongoing motion also affect where the gripper moves. For the next $H$ actions $U=(u_0,\ldots,u_{H-1})$, the shield predicts end-effector poses using executed-motion history $\mathcal H_t$ and a short-horizon response model. Translation is approximated by critically damped dynamics,
\begin{equation}
    \ddot p = \omega_n^2(p^{\mathrm{goal}}-p)-2\omega_n\dot p,
    \label{eq:rp1}
\end{equation}
 where $p$ denotes end-effector position, with velocity $\dot p$ and acceleration $\ddot p$. 
The controller target \(p^{\mathrm{goal}}\) is obtained by adding the scaled translational component of the VLA-predicted action to the end-effector position at the start of the control step. The natural frequency $\omega_n$ sets the response rate. Starting from the observed motion state, we predict position and, through an analogous rotation model, orientation over the action chunk. The shield then propagates the gripper and any inferred carried-object geometry with these poses, so the next stage checks distances from the moving bodies to obstacles. 

Action filtering uses these predicted body positions to decide how much of the action chunk can execute. At step $k$, the shield computes the distance $d_{a,k}$ between each body $a$ (the gripper or carried object) and its obstacle set. It releases only the actions before the first predicted step at which any distance reaches or falls below its threshold $\delta_a$. The resulting prefix length is
\begin{equation}
g=\min\Bigl(
\{k\in\{0,\ldots,H-1\}:\min_a(d_{a,k}-\delta_a)\leq 0\}
\cup\{H\}\Bigr).
\label{eq:prefix}
\end{equation}
When $g>0$, it releases $(u_0,\ldots,u_{g-1})$, and $g=H$ permits the entire checked chunk. When $g=0$, it withholds the candidate actions and commands a short retreat upward and away from the forbidden point nearest the end effector. It then reassesses using fresh observations. History is updated only from actions actually executed and their observed motion, so withheld actions are not treated as changes to the physical state.

\subsubsection{Reactive recovery with reusable skills}
\label{sec:method-reactive}

Robust task completion requires recovering from disturbances that have already interrupted manipulation. At this point, filtering the next action may reduce further contact risk but cannot by itself restore task progress. We therefore implement \emph{reactive recovery} with control-based skills that attempt to repair the physical state before the VLA retries the interrupted subtask. The \emph{recovery skill library} initially contains one shared base skill for all targets. When the Judge detects a subtask failure, the runtime suspends the VLA and invokes the applicable recovery behavior from the current library version, using retained revisions where available or base version otherwise.

The base skill proceeds through observation, retrieval, verification, and resumption. For a dropped object, it first moves the arm to an observation configuration and re-localizes the target. It generates grasp candidates, executes a grasp, then lifts and checks holding from the measured gripper width. A passing check allows the arm to carry the object to a stand-off pose near the destination, where a second check assesses whether the grasp was retained during transport. If this check passes, the VLA resumes the original instruction from that pose. If retrieval fails or the second holding check does not pass, the recovery attempt is marked as failed. The runtime then attempts to return to the available grasp candidate's pre-grasp pose, or to the recorded pre-failure pose if no candidate is available, and opens the gripper on arrival. Both paths continue in the current scene within the remaining episode budget, and the Judge assesses task completion separately from recovery and resumption.

\subsubsection{Refining recovery skills across attempts}
\label{sec:method-refinement}

Failures can recur when recovery repeats an unsuitable grasp configuration or an ineffective sequence of operations. Improving recovery can therefore require changing an operation's numerical settings or the procedure in which it is used. We use offline reflection to refine the current skill version from unresolved failures. Between evaluation rounds, reflectors inspect selected failed attempts using scene and target images, grasp-stage measurements, and the version used. Parameter memory (P) records changes to numerical settings, for example, a revised contact tool-center-point height is applied when constructing a grasp command. Strategy memory (S) records changes to the recovery control flow, such as adding retries or VLM-assisted re-grounding. With re-grounding enabled, the runtime segments current object candidates and asks a VLM to compare them with the target reference image. The selected mask is then passed to the grasp-generation routine.

After reflection produces proposed revisions, an admission check rejects unsupported formats and out-of-range numerical values. Accepted P/S updates are incorporated into the current recovery procedure and saved as the next library version. The following round evaluates that version, and its unresolved failures guide further reflection and refinement. Successive rounds build on the preceding version while the VLA weights remain fixed. Appendix~\ref{app:recovery-runtime-memory} details how the records are used, and Section~\ref{sec:exp-recovery} specifies failure selection and the two-round protocol.
\section{Experiments}
\label{sec:experiments}

\begingroup
\setlength{\textfloatsep}{6pt plus 2pt minus 2pt}
\setlength{\floatsep}{6pt plus 2pt minus 2pt}
\setlength{\intextsep}{6pt plus 2pt minus 2pt}

We evaluate the planning, shielding, and recovery mechanisms. Real-robot studies test their use in physical assistance (Fig.~\ref{fig:exp-personalization}); controlled simulation studies extend the evaluation across scenes, safety constraints, and disturbances (Fig.~\ref{fig:exp-simulation}).

\begin{figure}[!t]
\centering
\includegraphics[width=0.8\linewidth]{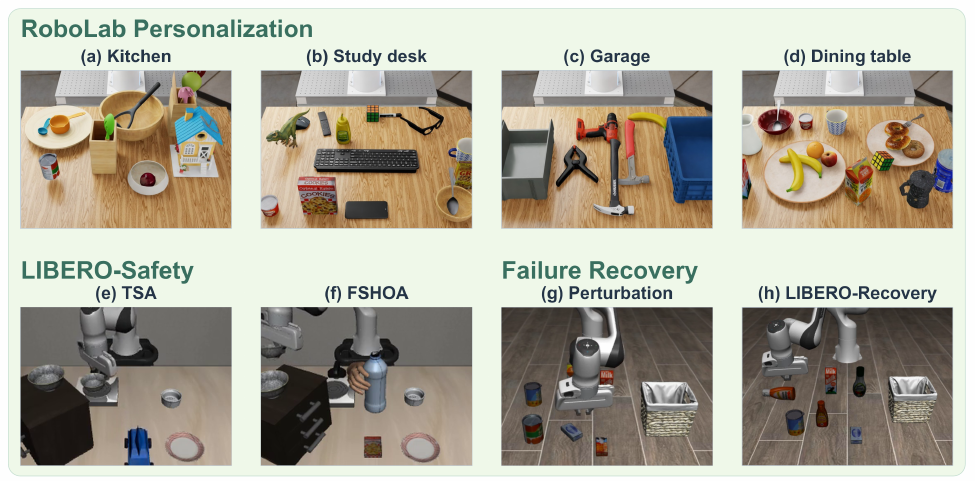}
\caption{\textbf{Simulation settings.} (a)--(d) RoboLab scenes for personalized planning and memory revision; (e)--(f) LIBERO-Safety tasks for proactive shielding; (g)--(h) controlled perturbations in LIBERO and recorded failure states in LIBERO-Recovery.}
\label{fig:exp-simulation}
\end{figure}

\begin{table}[!htbp]
\centering\small
\setlength{\tabcolsep}{3pt}
\caption{Real-robot results. Planning rows count correct subtasks; execution rows count completed episodes.}
\label{tab:realbot}
\label{tab:realbot-recovery}
\begin{minipage}[t]{0.61\linewidth}
\centering
\textbf{Planning loop}\par\smallskip
\begin{tabularx}{\linewidth}{@{}>{\raggedright\arraybackslash}p{0.35\linewidth}>{\raggedright\arraybackslash}Xr@{}}
\toprule
Study & Condition & Result \\
\midrule
\multirow{3}{=}{Sandwich Preparation} & Textual & 24 / 24 \\
 & Memory & 24 / 24 \\
 & Preference update & 20 / 24 \\
\addlinespace[3pt]
\multirow{3}{=}{Activity-dependent Packing} & Memory & 16 / 16 \\
 & No-habit & 16 / 20 \\
 & Isolation & 16 / 18 \\
\addlinespace[3pt]
\multirow{3}{=}{Covered ingredient, scene memory} & Write and retrieve & 12 / 12 \\
 & Search without memory & 8 / 10 \\
 & Revise stale memory & 9 / 12 \\
\bottomrule
\end{tabularx}
\end{minipage}\hfill
\begin{minipage}[t]{0.36\linewidth}
\centering
\textbf{Execution loop}\par\smallskip
\begin{tabularx}{\linewidth}{@{}>{\raggedright\arraybackslash}Xr@{}}
\toprule
Condition & Result \\
\midrule
\multicolumn{2}{@{}l}{\emph{Obstacle avoidance}} \\
Raw policy & 0 / 20 \\
Ours, contact-free & 20 / 20 \\
\midrule
\multicolumn{2}{@{}l}{\emph{Recovery after object removal}} \\
RAW & 3 / 20 \\
BASE ($z$ 0.00, 1 attempt) & 9 / 20 \\
FULL ($z$ $-0.02$, 3 attempts) & 18 / 20 \\
\bottomrule
\end{tabularx}
\end{minipage}
\end{table}

\subsection{Personalization across Interactions and Scenes}
\label{sec:exp-personalization}

We test whether persistent context supports appropriate task decisions, remains useful after updates, and reduces scene search. Real-robot studies report subtask dispatch accuracy: the fraction of dispatched subtasks judged correct under task requirements and applicable user preferences (Table~\ref{tab:realbot}). RoboLab complements these studies with controlled endpoint selection and memory revision.

\leadpara{Sandwich preparation: retaining and revising preferences.}
This experiment tests whether user memory preserves user preferences across conversations and supports plan changes when those preferences are revised. The robot prepares sandwiches from toast, lettuce, and ham across four ingredient layouts. We compare three conditions: \emph{Textual} specifies how to make the sandwich directly in the query; \emph{Memory} introduces the user's preference in an earlier conversation, restarts the program, and then asks only ``Make me a sandwich''; \emph{Preference update} uses another conversation to revise the user's dietary preference before the same request. Conversations also contain unrelated remarks, requiring only relevant preferences to be retained. Dispatch accuracy is 100\% (24/24) for both Textual and Memory, and 83.3\% (20/24) after updates. These results show that the system retains user preferences across conversations and follows revised preferences in 20 of 24 dispatched subtasks.

\begin{figure}[!t]
\centering
\includegraphics[width=0.8\linewidth]{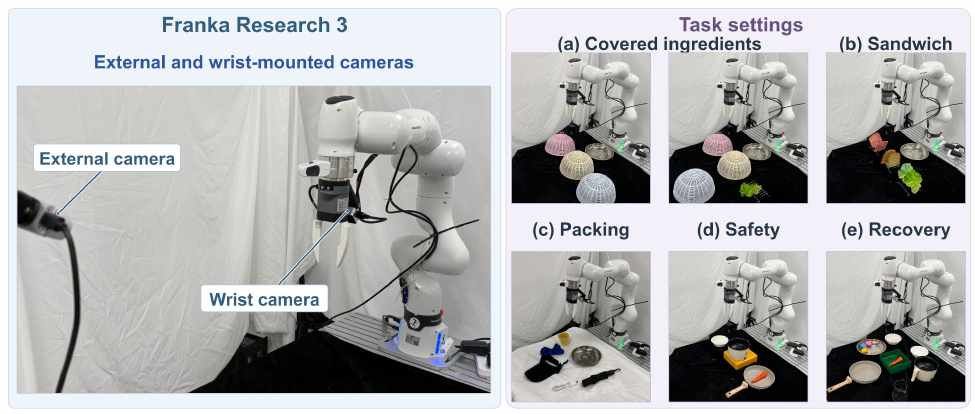}
\caption{\textbf{Real-robot settings.} A Franka Research 3 with external and wrist-mounted cameras. (a) Covered-ingredient retrieval before and after lid opening; (b) sandwich preparation; (c) activity-dependent packing; (d) safe execution; (e) failure recovery.}
\label{fig:exp-personalization}
\end{figure}

\leadpara{Activity-dependent packing: combining profiles, intent, and online information.}
This experiment tests whether the planner can model the user's profile with more complex information to make personalized decisions in a multi-user setting. The robot selects from a cap, cup, umbrella, charger, and towel for different activities such as running, visiting the gym, going to the park, or commuting to work. The planner queries the current weather at the user's residence through online search and combines it with the requested activity and personal habits. Two users have distinct residences and habits, including
preferences that differ from generic packing choices. \emph{Memory} supplies the user's profile through conversation before restarting the program; \emph{No-habit} supplies the same information without personal habits and relies on commonsense reasoning for planning; \emph{Isolation} registers both profiles before one user's request to test profile isolation. Required and forbidden items are judged against recorded weather and preferences. Dispatch accuracies are 100\% (16/16), 80.0\% (16/20), and 88.9\% (16/18), respectively. These results support combining persistent user profiles,
current intent, and online information for personalized
decisions in a multi-user setting.

\leadpara{Covered ingredients: retrieving and correcting scene memory.}
This experiment tests whether scene memory guides the search for hidden targets and is revised when observations contradict remembered locations. Lettuce, toast, and ham are hidden beneath colored lids, and the robot is asked to transfer a named ingredient to a plate. \emph{Write and retrieve} provides its location before a restart, followed by a request without a location hint. \emph{Search without memory} requires successive lid opening and observation before finding the target. \emph{Revise stale memory} swaps ingredients after memory formation to test if memory can be revised. Dispatch accuracies are 100\% (12/12), 80.0\% (8/10), and 75.0\% (9/12), respectively. Remembered locations support retrieval, while fresh observations can correct the stale context used for planning. Recorded failure cases include misidentification and omitted subtasks.

\leadpara{Multimodal requests and memory revision across scenes.}
We test whether memory enables efficient scene selection for named and visually specified personal objects across four
endpoints in RoboLab~\cite{yang2026robolab}: garage, dining table, kitchen, and study desk. The planner must select where to act for a target specified by name or photograph, allowing users to refer to personal objects without a precise verbal description. With no initial scene view, the planner requests endpoint images or consults stored scene memory. Each text/image condition contains 88 requests. Image queries never name the object: without memory, the photograph accompanies the instruction \emph{go pick up this}; with memory, \emph{go pick up mine} refers to a photograph supplied in an earlier conversation as \emph{this is mine}. Memory raises correct selections in both query modes and reduces latency (Table~\ref{tab:g2}a). These configurations support both named and remembered visual references while reducing the search required to select a scene.

\begin{table}[!htbp]
\centering\small
\setlength{\tabcolsep}{3pt}
\caption{RoboLab planning without physical execution. (a) Static scenes: 88 requests per condition; latency is the median time to dispatch. (b) Relocated targets: 88 request pairs per query type. Memory corrected requires both writing the new location and removing the stale association.}
\label{tab:g2}
\begin{minipage}[t]{0.48\linewidth}
\centering
\textbf{(a) Query support and efficiency}\par\smallskip
\begin{tabular*}{\linewidth}{@{\extracolsep{\fill}}lrr@{}}
\toprule
Query & \shortstack[r]{Correct\\/ 88} & \shortstack[r]{Time\\(s)}\\
\midrule
Text & 77 & 28.6\\
Image & 77 & 34.6\\
Text + memory & 88 & 7.0\\
Image + memory & 88 & 7.1\\
\bottomrule
\end{tabular*}
\end{minipage}\hfill
\begin{minipage}[t]{0.48\linewidth}
\centering
\textbf{(b) Closed-loop scene-memory revision}\par\smallskip
\begin{tabular*}{\linewidth}{@{\extracolsep{\fill}}lrr@{}}
\toprule
Query & \shortstack[r]{Memory\\corrected} & \shortstack[r]{Follow-up\\dispatch}\\
\midrule
Text & 76 / 88 & 83 / 88\\
Image & 77 / 88 & 85 / 88\\
\bottomrule
\end{tabular*}
\end{minipage}
\end{table}

We additionally conduct a revision experiment to test whether new observations correct persistent scene records and whether later requests can use the revised memory for endpoint selection. We relocate targets while keeping their old locations in memory. A valid memory correction must write the new location and remove the stale association. Corrections succeed in 76/88 text and 77/88 image trials; follow-up dispatch is correct in 83/88 and 85/88, respectively (Table~\ref{tab:g2}b). The two measures support observation-driven revision and later reuse of scene knowledge.

\leadpara{Analysis.}
The above studies evaluate the lifecycle of personalized context: acquiring requirements, retaining them across interactions, and revising them after change. The results support connecting memory to planning and observation, rather than treating stored profiles and locations as permanently valid instructions.

\subsection{Safety during Execution}
\label{sec:safety}

\leadpara{Geometric checks beyond safety fine-tuning.}
We test whether runtime shielding reduces unsafe contact beyond improvements from training with safe demonstrations. LIBERO-Safety~\cite{cui2026libero} provides tabletop spatial avoidance (TSA) and free-space hand-object avoidance (FSHOA), with 450 episodes per suite. We evaluate the original and safety-finetuned (SFT) $\pi_{0.5}$, with and without our shield, alongside adapted VLSA~\cite{hu2025vlsa}, and extend the comparison to OpenVLA-OFT~\cite{kim2025fine} and VLAct~\cite{yang2026beyond} to test transfer across policy backbones (Table~\ref{tab:safety}). The official SFT $\pi_{0.5}$ checkpoint was released by~\cite{cui2026libero} and was trained on 19,664 safe demonstrations for collision-free execution. We report violation-free task success rate (VF) and violation rate (VIOL.).

\begin{table}[htbp]
\centering
\small
\setlength{\tabcolsep}{3.5pt}
\caption{Physical safety by suite and difficulty on LIBERO-Safety. Each entry is VF $\uparrow$ / Viol. $\downarrow$ (\%): violation-free task success and episodes with at least one constraint violation. Each suite--level cell contains 150 episodes.}
\label{tab:safety}
\begin{tabular}{l*{7}{r}}
\toprule
 & \multicolumn{3}{c}{TSA} & \multicolumn{3}{c}{FSHOA} & Overall \\
\cmidrule(lr){2-4} \cmidrule(lr){5-7}
Method & L0 & L1 & L2 & L0 & L1 & L2 & All \\
\midrule
VLAct~\cite{yang2026beyond} & 0.0 / 93.3 & 20.7 / 30.7 & 0.0 / 85.3 & 0.0 / 76.0 & 5.3 / 47.3 & 0.0 / 83.3 & 4.3 / 69.3 \\
VLAct + VLSA & 17.3 / 66.7 & 20.0 / 20.0 & 10.7 / 58.7 & 1.3 / 52.0 & 15.3 / 24.0 & 13.3 / 47.3 & 13.0 / 44.8 \\
VLAct + Ours & 20.0 / 25.3 & 20.0 / 24.7 & 20.0 / 28.0 & 11.3 / 26.0 & 10.7 / 15.3 & 1.3 / 37.3 & \textbf{13.9} / \textbf{26.1} \\
\midrule
OpenVLA-OFT~\cite{kim2025fine} & 0.0 / 67.3 & 20.7 / 38.7 & 6.0 / 61.3 & 0.0 / 78.7 & 14.7 / 52.0 & 0.0 / 70.7 & 6.9 / 61.4 \\
OpenVLA-OFT + VLSA & 12.7 / 47.3 & 19.3 / 22.7 & 6.7 / 45.3 & 0.7 / 46.7 & 18.0 / 46.0 & 16.7 / 44.0 & 12.3 / 42.0 \\
OpenVLA-OFT + Ours & 20.0 / 16.0 & 19.3 / 8.0 & 19.3 / 12.0 & 7.3 / 20.7 & 18.0 / 10.0 & 4.7 / 36.7 & \textbf{14.8} / \textbf{17.2} \\
\midrule
$\pi_{0.5}$~\cite{intelligence2025pi_} & 0.0 / 84.7 & 26.7 / 34.7 & 0.0 / 84.0 & 1.3 / 72.0 & 12.0 / 66.0 & 0.0 / 85.3 & 6.7 / 71.1 \\
$\pi_{0.5}$ + VLSA & 8.0 / 66.7 & 26.7 / 10.7 & 5.3 / 65.3 & 2.7 / 52.0 & 18.7 / 45.3 & 14.7 / 58.7 & 12.7 / 49.8 \\
$\pi_{0.5}$ + Ours & 18.7 / 18.7 & 24.0 / 17.3 & 20.0 / 16.0 & 10.7 / 27.3 & 12.0 / 18.0 & 2.0 / 38.7 & \textbf{14.6} / \textbf{22.7} \\
\midrule
$\pi_{0.5}$-SFT~\cite{cui2026libero} & 17.3 / 69.3 & 40.7 / 23.3 & 24.0 / 61.3 & 20.0 / 76.0 & 35.3 / 4.7 & 14.7 / 80.7 & 25.3 / 52.6 \\
$\pi_{0.5}$-SFT + VLSA & 15.3 / 56.7 & 38.7 / 16.0 & 19.3 / 42.0 & 19.3 / 51.3 & 30.7 / 6.0 & 29.3 / 54.0 & 25.4 / 37.7 \\
$\pi_{0.5}$-SFT + Ours & 20.0 / 13.3 & 40.0 / 6.0 & 23.3 / 4.7 & 18.7 / 4.7 & 35.3 / 4.7 & 16.0 / 3.3 & \textbf{25.6} / \textbf{6.1} \\
\bottomrule
\end{tabular}
\end{table}

Despite safety fine-tuning, 52.6\% of episodes incur violations. Shielding reduces this to 6.1\%, compared with 37.7\% for adapted VLSA, while VF remains similar at 25.3\% without shielding and 25.6\% with it. For the original checkpoint, violations decrease from 71.1\% to 22.7\%, compared with 49.8\% for adapted VLSA, while VF increases from 6.7\% to 14.6\%. The same shield also reduces violations with OpenVLA-OFT, from 42.0\% under VLSA to 17.2\%, while increasing VF from 12.3\% to 14.8\%. On VLAct, our shield reduces violations from 44.8\% under VLSA to 26.1\%, with a smaller VF increase from 13.0\% to 13.9\%, supporting cross-backbone violation reduction.

For deployment cost, measured mean per-invocation times are 21.06\,ms for geometric filtering, 47.82\,ms for VLA inference, and 170.82\,ms for perception. Perception dominates these measurements, while filtering contributes a smaller computational component.

\leadpara{Real-world safe execution.}
The physical study tests whether the proactive shield reduces contact during real-robot execution when an obstacle is introduced. The real-robot policy is trained on carrot relocation without obstacles. With an inserted obstacle, contact-free completion increases from 0/20 for the raw policy to 20/20 with shielding (Table~\ref{tab:realbot}), showing repeatable intervention against collision hazards. 
 


\begin{figure}[!t]
\centering
\includegraphics[width=0.8\linewidth]{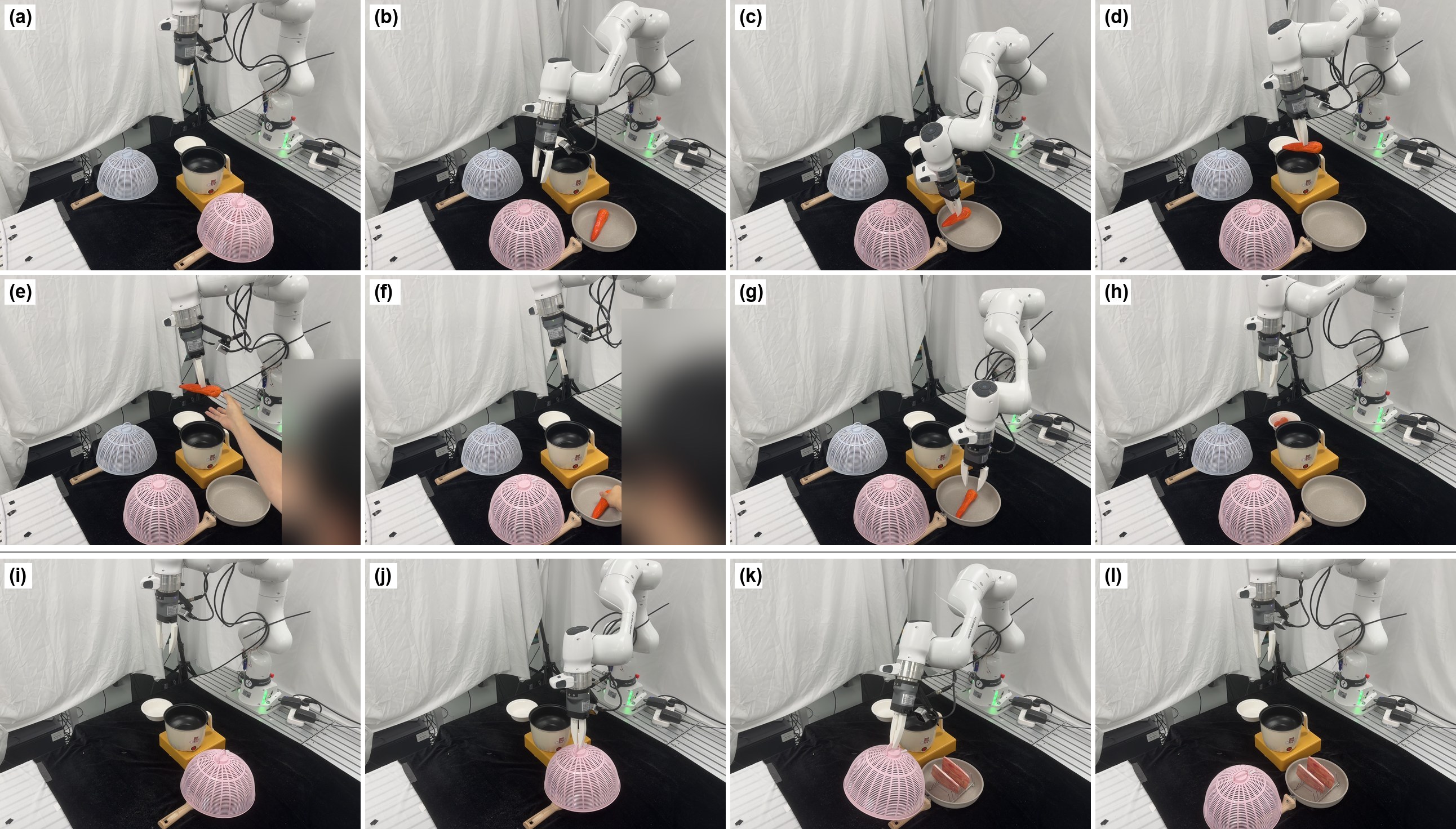}
\caption{\textbf{Integrated food-serving runs.} (a)--(d) Memory-guided lid opening and shielded transport. (e)--(h) Recovery restores a lost grasp before task completion. (i)--(l) Revealing meat triggers a scene-memory update and refusal under the vegetarian preference.}
\label{fig:integrated}
\end{figure}

\subsection{Recovery and Improvement from Failure Experience}
\label{sec:exp-recovery}

\leadpara{Recovery from perturbation}
This experiment tests whether the recovery skill restores task progress after disturbance and whether using successively refined skills in reflection-and-retry rounds increases cumulative completion. The disturbance-response results
(Fig.~\ref{fig:disturbance-response}) motivate control-based recovery when perturbations undermine VLA execution. 

The experiments are conducted on the LIBERO-Object task suite with a 12\,cm positive-x hold-loss perturbation. R0 uses the shared base recovery skill with empty strategy (S) and parameter (P) memories. After each round, reflection uses up to three failed episodes with recovery evidence per object. Baselines and our reflection process use Qwen3-VL-235B-A22B-Instruct. Harness VLA uses the same Qwen3-VL-4B-OFT VLA checkpoints as our system, while CaP-X does not require a VLA. 

\begin{table}[!htbp]
\begin{minipage}[t]{0.48\linewidth}
\centering\small
\setlength{\tabcolsep}{3pt}
\caption{Recovery on 200 LIBERO episodes with perturbation. R1/R2 include additional reflection and retries. API tokens accumulate execution and reflection calls per original episode.}
\label{tab:recovery}
\label{tab:reflection-rounds}
\begin{tabularx}{\linewidth}{@{}>{\raggedright\arraybackslash}Xrr@{}}
\toprule
System / Stage & \shortstack[r]{Cumulative\\/ 200 (\%)} & \shortstack[r]{API tokens\\per episode}\\
\midrule
Qwen3-VL-OFT    & 26 (13.0) & ---\\
CaP-X~\cite{fu2026cap}  & 10 (5.0) & $ 1{,}124{,}819$\\
Harness VLA~\cite{zhang2026harness}  & 101 (50.5) & $ 3{,}829{,}605$\\
\midrule
\sysname{} R0: base skill  & 58 (29.0) & $ 21{,}371$\\
\sysname{} R1: one reflection  & 141 (70.5) & $34{,}393$\\
\sysname{} R2: two reflections  & \textbf{160 (80.0)} & $41{,}109$\\
\bottomrule
\end{tabularx}
\end{minipage}\hfill
\begin{minipage}[t]{0.48\linewidth}
\centering\small
\setlength{\tabcolsep}{3pt}
\caption{Recovery on 770 LIBERO-Recovery Object cases. R1/R2 include additional reflection and retries. API tokens accumulate execution and reflection calls per original episode.}
\label{tab:recorded-recovery}
\begin{tabularx}{\linewidth}{@{}>{\raggedright\arraybackslash}Xrr@{}}
\toprule
System / Stage & \shortstack[r]{Cumulative\\/ 770 (\%)} & \shortstack[r]{API tokens\\per episode}\\
\midrule
Pure VLA & 247 (32.08) & 0\\
\sysname{} R0: base skill & 385 (50.00) & 2,674\\
\sysname{} R1: one reflection & 569 (73.90) & $5{,}492$\\
\sysname{} R2: two reflections & \textbf{586 (76.10)} & $7{,}159$\\
\bottomrule
\end{tabularx}
\end{minipage}
\end{table}

With base recovery, R0 completes 58/200 episodes (29.0\%). R1 resolves 83/142 previous failures, raising cumulative completion to 70.5\%; R2 resolves another 19/59, reaching 80.0\% (Table~\ref{tab:recovery}). The second reflection-and-retry round extends recovery beyond the first revision, and the full iterative system achieves the highest completion rate. 

To assess the VLM-query cost of recovery-focused refinement, we measure billed API tokens including both reflection rounds and all retries, amortized over the original episodes. Our recorded usage is 41,109 tokens per episode, including 2,609 for reflection, compared with the recorded baseline totals in Table~\ref{tab:recovery}. Higher cumulative completion together with lower recorded VLM-token usage supports recovery-focused refinement as a more efficient method.

\leadpara{Recovery from recorded failure states.}
\label{sec:recorded-recovery}
We further evaluate recovery from natural execution-induced failure states using LIBERO-Recovery~\cite{liu2026libero}. The evaluation covers 77 Object scenes with five initially closed-gripper and five initially open-gripper trials each, yielding 770 cases without additional runtime disturbances. Since failure states are supplied, we disable the Judge and invoke recovery after scene initialization. Pure VLA and R0 share the same Qwen3-VL-4B-OFT VLA checkpoint.

Base recovery raises completion from 32.08\% to 50.00\%, a gain of 17.92 percentage points under the same rollout budget. R1 resolves 184/385 previous failures, increasing cumulative completion to 73.90\%; R2 resolves another 17/201, reaching 76.10\% (Table~\ref{tab:recorded-recovery}). Final cumulative API usage is 7,158.77 tokens per original case, including reflection and retries. The above results support iterative recovery from recorded natural failure states.

\leadpara{Real-world recovery.}
This experiment tests whether base recovery and retained skill revisions improve real-robot task completion after object removal. An operator takes the lifted object and places it at one of five locations, with two orientations and two repetitions (20 episodes per condition). We compare the raw policy, base recovery, and a revised skill combining a $-0.02$\,m grasp-height offset with up to three attempts and renewed perception. Task completion rises from 3/20 to 9/20 and 18/20, respectively (Table~\ref{tab:realbot}). External repair improves completion over the raw policy, and retained revisions improve it further across the tested placements.

\leadpara{Analysis.}
The execution studies support complementary interventions around a frozen VLA: shielding reduces contact violations, while recovery restores progress after disruption. Reflection-and-retry rounds extend cumulative completion, and revisions improve real-robot repair. Together, the two paths combine immediate protection with reusable failure recovery skills.

\subsection{Integrated Assistance on the Real Robot}
\label{sec:exp-realbot}
\label{sec:exp-integrated}

\leadpara{Setup.}
This study examines whether personalized planning, proactive shielding, and reactive recovery operate as one system in a long-horizon assistance task, where each component acts on what the others produce. In the food-serving scenario of Fig.~\ref{fig:teaser}, a user with a stored vegetarian preference asks the robot to serve lunch from a covered pan. The robot must look under the lid, decide whether the food may be served, and carry it past an obstacle while an operator disturbs the transport. A second trial places meat under the lid while scene memory still records a carrot. All real-robot experiments use a Qwen3-VL-4B-OFT policy~\cite{community2026starvla}; the policy for this task was trained on lid opening and food pickup without obstacles.

\leadpara{Results.}
In the carrot run (Fig.~\ref{fig:integrated}a--h), the robot uncovers the pan guided by scene memory and starts the transfer. The shield avoids potential collisions with the rice cooker obstacle during transport. After the carrot is dropped into the pan as a perturbation, the VLA is not able to pick it up again. The Judge triggers the recovery procedure, which re-grasps the carrot and hands control back to the VLA to place it in the bowl. In the meat run (Fig.~\ref{fig:integrated}i--l), the robot uncovers the pan and finds a contradiction with scene memory, then replaces the stale carrot entry with the observed meat entry. The system then declines to serve anything as the food currently on the table does not meet the user's dietary requirements. The accompanying video shows both runs.

\leadpara{Analysis.}
These runs illustrate how \sysname{} preserves the user's intent while adapting both the plan and its execution. Scene memory guides where to look, while fresh visual evidence and the stored dietary preference determine whether serving should proceed. In the carrot run, the obstacle and lost grasp disrupt execution without invalidating the goal: shielding adjusts the transfer motion, and Judge-triggered recovery restores the grasp before returning control to the VLA to finish serving. In the meat run, the newly revealed food instead invalidates the planned transfer under the same dietary constraint. Visual feedback therefore changes the planning decision and corrects the persistent scene record, rather than merely checking whether a manipulation succeeded. The benefit of the dual-loop design is this distinction between repairing an interrupted, appropriate task and stopping a task that new evidence makes inappropriate. Together, the runs provide evidence that contextual decisions, action-level protection, and physical repair can support a common assistance goal around a frozen VLA.

\endgroup

\FloatBarrier 
\section{Conclusion}
\label{sec:conclusion}

We presented \sysname{}, a dual-loop harness for personalized and reliable robot assistance. The planning loop maintains revisable user and scene memories to guide task decisions across interactions, query modalities, and execution endpoints. The execution loop combines proactive geometric shielding with reactive recovery, using offline reflection to refine recovery skills while keeping the underlying VLA policy frozen. Real-robot and simulation studies evaluate these complementary capabilities, from preference-conditioned assistance and scene-memory revision to runtime protection and recovery after disturbance. Together, they show that persistent context and physical feedback extend robot assistance beyond isolated task execution. Future work will evaluate longer-term adaptation and the transfer of recovery experience across tasks, while improving the balance between protective intervention and task completion.

\restoregeometry
\bibliographystyle{unsrt}
\bibliography{references}

\clearpage
\appendix
%

\providecommand{\authorcheck}[1]{\textcolor{red}{[Author check: #1]}}

\begin{center}
{\sffamily\Large\bfseries\color{darkblue} Appendix}
\end{center}


\section{System Implementation and Runtime Coordination}
\label{app:system}

This appendix specifies how control passes between the planner, the VLA, the Judge, the shield, and recovery while a request is handled (Fig.~\ref{fig:architecture}). 
Each planner output is a \emph{plan}: a goal and a set of subtasks, each of which sends an instruction to an endpoint. Invalid plans, including those that omit a required \texttt{update\_scene} (Appendix~\ref{app:scene-update}), are returned to the planner for repair. Algorithm~\ref{alg:app-runtime} summarizes one request. A subtask without dependencies starts immediately, and a subtask that depends on earlier subtasks starts in sequence.
\label{app:interfaces}
\label{app:control-flow}

\begin{algorithm}[htbp]
\caption{Handling one request in the dual-loop harness}
\label{alg:app-runtime}
\small
\begin{algorithmic}[1]
\Require request $c$; user memory $M^u$; scene memory $M^s$; endpoints $\mathcal E$; skill library $\mathcal L$
\State $x \gets \textsc{BuildContext}(c, M^u, M^s, \mathcal E)$
\State $P \gets \textsc{Planner}(x)$, repaired until valid
\ForAll{subtask $\ell_i \in P$ whose dependencies succeeded}
  \If{$\ell_i$ writes memory} \State write the record to $M^s$ or $M^u$ \Comment{\texttt{update\_scene}, \texttt{update\_user\_memory}}
  \ElsIf{$\ell_i$ calls a non-physical skill} \State run the skill; append its result to $x$ \Comment{e.g., web query, scene check}
  \Else \Comment{physical skill executed by the VLA: manipulation or observation step}
    \State dispatch $\ell_i$ to the VLA; start the Judge
    \While{the job is running}
      \State execute $\textsc{Shield}(\pi_\theta(o_t, \ell_i))$ \Comment{Section~\ref{sec:method-proactive}}
      \If{Judge reports \emph{success}} mark $\ell_i$ succeeded; append what it revealed to $x$
      \ElsIf{Judge reports \emph{failed}}
        \State suspend the VLA; run the active recovery skill \Comment{Section~\ref{sec:method-reactive}}
        \State resume the VLA on the \emph{original} $\ell_i$; restart the Judge
      \EndIf
    \EndWhile
  \EndIf
  \If{$\ell_i$ revealed new scene contents} replan; the next plan must record them with \texttt{update\_scene} \EndIf
\EndFor
\State on termination: $M^u \gets \textsc{Summarize}(\text{conversation}, M^u)$
\end{algorithmic}
\end{algorithm}

\section{Persistent User and Scene Memory}
\label{app:memory}

This appendix gives the storage format of both memories, their update paths, and logged examples showing how a revision changes the next plan. 

\subsection{Storage and Planner Injection}
\label{app:memory-schema}

Both memories are Markdown files of \texttt{\#\# <section>} blocks with \texttt{- key: value} lines (Table~\ref{tab:app-memory-schema}). Both are rendered into the planner context on every request. All user files are rendered under per-user headings, and the planner prompt instructs it to apply only the requesting user's facts. 

\begin{table}[htbp]
\centering\small
\setlength{\tabcolsep}{4pt}
\caption{Memory schemas.}
\label{tab:app-memory-schema}
\begin{tabularx}{\linewidth}{>{\raggedright\arraybackslash}p{2.0cm}>{\raggedright\arraybackslash}X>{\raggedright\arraybackslash}X}
\toprule
 & User memory $M^u$ & Scene memory $M^s$ \\
\midrule
Unit & One file per user; one section per category & One file; one section per place, named by the planner \\
Content & Facts in \texttt{profile:*} (identity, habits, preferences), \texttt{profile:avoid:*} (standing prohibitions), \texttt{entity:*} (owned objects, including\ \texttt{reference\_image}) & \texttt{- objects:} and \texttt{- description:} lines \\
Example & \texttt{\#\# profile:eating\_habit} / \texttt{- diet: vegetarian; does not eat meat} & \texttt{\#\# frying\_pan\_under\_pink\_lid} / \texttt{- objects: carrot} \\
Evidence & Dialogue & Observations \\
Writers & Summarizer; planner via \texttt{update\_user\_memory} & Planner via \texttt{update\_scene} only \\
Updates & Latest value per key; retraction by \texttt{delete} & Section replaced; a stale entry is retracted by rewriting it \\
\bottomrule
\end{tabularx}
\end{table}

\subsection{User Memory from Dialogue}
\label{app:user-memory}

User memory has two write paths. During a turn, the planner can record a preference the user states explicitly. In addition, after each plan finishes, a background summarizer reviews the recent conversation and writes durable facts that are not yet stored. Both follow the same rule for separating durable facts from temporary statements. A new value for a key replaces the old one, and a superseded preference is retracted.

\begin{prompt}{Durability rule in the summarizer prompt (excerpt, real-robot development branch)}
\begin{VerbatimWrap}
DURABILITY FILTER (the most important rule):
- Only write facts that will still be true in future sessions. Evidence must
  be a lasting preference/attribute explicitly stated, or clearly meant as a
  standing instruction ("always cook 2 eggs from now on").
- Do NOT write one-off / session-scoped wishes ("I don't want meat today",
  "no salt this time") or volatile data (weather, time, robot/task state).
- Negatives use the same filter: profile:avoid:* only for a STANDING prohibition,
  never one-off negatives.
\end{VerbatimWrap}
\end{prompt}

For example, when Bob says ``I take my charger with me everywhere, even to the gym. Also, I need to leave by five today.'', the profile stores \texttt{profile:habit/charger} and nothing about leaving at five. 

We show an example below to demonstrate how preference change is reflected in the user profile. In the sandwich experiment, the same request yields different plans before and after the stored diet changes. 

\begin{prompt}{Sandwich demonstration: plans under preference change}
\begin{VerbatimWrap}
# Profile: vegetarian
CarryOn> Make me a sandwich.
[plan] I'll make you a vegetarian sandwich with toast and lettuce.
  - t1: toast / t2: lettuce / t3: toast

# Profile after update: diet: non-vegetarian; prefers ham and lettuce in sandwiches
CarryOn> Make me a sandwich.
[plan] I'll make you a sandwich with ham and lettuce using the available ingredients.
  - t1: toast / t2: ham / t3: lettuce / t4: toast
\end{VerbatimWrap}
\end{prompt}

\subsection{Scene Memory from Observation}
\label{app:scene-update}

An observation that reveals new contents creates an outstanding record, and the validator rejects the next plan unless it contains an \texttt{update\_scene} subtask. The RoboLab example below shows a stale location being revised and reused. Memory lists a Rubik's cube at the garage bench, but the cube is on the dining table. The first request checks the remembered station, rewrites it, finds the cube elsewhere, and records the new location; the follow-up dispatches directly. 

\begin{prompt}{RoboLab relocated-target pair (text query)}
\begin{VerbatimWrap}
# Scene memory before:  ## franka_garage_bench / - objects: rubiks cube
[plan] I'll check the garage bench for the Rubik's cube first.
  - t1 @ check_endpoint -> {"present": false, ...}
[plan] I'll update the record for the garage bench and then check the dining table.
  - t1 @ update_scene: Record that the garage bench currently holds no Rubik's cube.
  - t2 @ check_endpoint -> {"present": true, "confidence": 0.95, "scene_update_required": true,
                             "revealed_place": "franka_dining_table", ...}
[plan] I'll record that the Rubik's cube is on the dining table and then pick it up.
  - t1 @ update_scene: Record the presence of the Rubik's cube at the dining table.
  - t2 @ franka_dining_table: Pick up the Rubik's cube from the dining table.
# Scene memory after:   ## franka_dining_table / - objects: rubiks cube
#                       ## franka_garage_bench / - objects:
# Follow-up request (new process, memory carried over):
[plan] I will pick up the Rubik's cube from the dining table.
\end{VerbatimWrap}
\end{prompt}

\subsection{Remembered Reference Images}
\label{app:memory-images}

A reference image is stored as an entity attribute: the file is copied into the memory store, and a \texttt{reference\_image} fact pointing to it is written under \texttt{entity:<name>}, either by the summarizer when the model tags an entity fact with an attached image or by an explicit \texttt{/remember\_image <path> <entity>} command. A later request can then refer to the entity (``go pick up mine''), and the stored image is available to the planner through the user-memory block. 


\section{Proactive Shielding}
\label{app:shield}
\label{app:shield-grounding}
\label{app:shield-protocol}
\label{app:setup-safety}

The LIBERO-Safety scene builder assigns task roles to observed geometry before applying contact rules. It receives target and destination names from the instruction. Text-prompted SAM3 segmentation associates these names with image masks; the corresponding depth pixels are back-projected into the world frame. When text grounding returns no object mask, the scene builder can use a VLM point prompt followed by point-prompted segmentation. The self-mask removes robot and mount pixels. The target's source support is obtained as a local geometric patch beneath the grounded target.

\begin{figure}[htbp]
\centering
\includegraphics[width=0.8\linewidth]{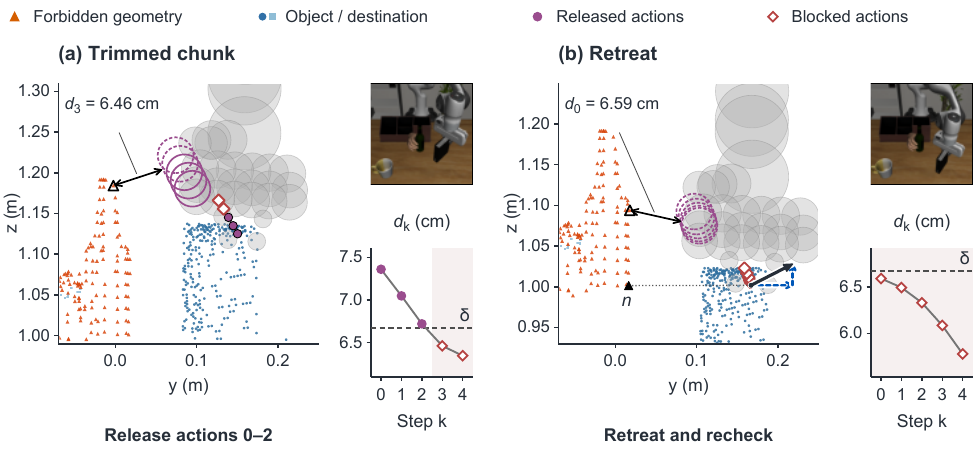}
\caption{\textbf{Geometric action filtering and retreat.} Each example shows a geometric side view (left), the observed scene (upper right), and predicted gripper--obstacle distances (lower right); gray spheres depict current robot geometry and purple outlines depict predicted gripper geometry. (a) Actions before the first violation of the dashed distance threshold are released; (b) if the first action is blocked, the robot retreats and rechecks. Open triangles mark obstacle points nearest the predicted gripper surface for filtering, whereas the filled triangle $n$ is nearest the current end-effector reference point and determines retreat. The two thin dashed arrows in (b), arranged head-to-tail, show the component away from $n$ and the added upward bias; the thick solid arrow shows their resulting retreat direction.}
\label{fig:app-shield-geometry}
\end{figure}

The contact rule exempts intended target, source-support, and destination contact for the bodies to which those permissions apply. Remaining scene obstacles are checked geometrically. Gripper geometry is propagated from each predicted end-effector pose. A holding-state latch uses measured gripper width and the executed gripper command to determine when carried-object geometry should be included. The inferred held-object point set is transformed using its attachment pose, so carried-object contact is checked in addition to gripper contact. If the first candidate action is blocked, the controller executes a three-control-step retreat with an away-from-obstacle component and an upward component. The retreat direction uses the obstacle point nearest the current end-effector reference point, and the filtering distance uses the surfaces of the predicted bodies, as distinguished in Fig.~\ref{fig:app-shield-geometry}.

Table~\ref{tab:app-shield-params} gives the parameters of the response model and decision rule on LIBERO-Safety. Figure~\ref{fig:app-shield-geometry} illustrates action filtering and retreat.

\begin{table}[htbp]
\centering\small
\setlength{\tabcolsep}{5pt}
\caption{Shield parameters on LIBERO-Safety.}
\label{tab:app-shield-params}
\begin{tabularx}{\linewidth}{>{\raggedright\arraybackslash}p{4.6cm}>{\raggedright\arraybackslash}X}
\toprule
Parameter & Value \\
\midrule
Controller & Operational-space pose control, $k_p=150$, damping ratio 1, control period 0.05\,s; $\omega_n=\sqrt{150}\approx12.25$\,s$^{-1}$ \\
Prediction horizon $H$ & 5 actions \\
Intervention threshold $\delta$ & 6.67\,cm (1.51\,cm margin $+$ largest displacement over 5 steps) \\
\bottomrule
\end{tabularx}
\end{table}

\section{Recovery Skills and Reflection}
\label{app:recovery}

This appendix describes additional details of the recovery and reflection mechanism.

\subsection{Base Recovery Skill}
\label{app:recovery-base}

The base skill attempts to restore the physical state needed to continue the interrupted subtask (Algorithm~\ref{alg:app-recovery}). After suspending the VLA, it moves the arm to an observation configuration, locates the target, and generates grasp candidates for retrieval. A check of measured gripper width after lifting determines whether to proceed with transport. If this check passes, the skill locates the destination, moves to a stand-off pose near it, and checks holding again. Passing the second check allows the VLA to resume the original instruction from that pose.

Recovery fails if the skill cannot locate the target or destination, its grasp attempts do not establish a hold, a required motion cannot be completed, or the second holding check does not pass. In these cases, the skill stops and the runtime attempts to return the arm to the available grasp candidate's pre-grasp pose, or to the recorded pre-failure pose if no candidate exists. It opens the gripper after arrival, then requests VLA resumption with the original instruction and remaining episode budget. Resumption is still requested if the fallback motion fails. The Judge continues to assess task completion; a successful recovery only enables the task to continue. 

\begin{algorithm}[htbp]
\caption{Base recovery skill (hold loss)}
\label{alg:app-recovery}
\small
\begin{algorithmic}[1]
\Require Judge \emph{failed} verdict for the current subtask
\State suspend the VLA; move to the initial configuration; open the gripper
\State ground the task target
\State build grasp candidates for the target \Comment{short axis, then long axis}
\For{each candidate}
  \State execute the grasp; lift and check holding from measured width
  \If{held} \textbf{break} \EndIf
\EndFor
\If{held}
  \State ground the destination; move to a pre-place stand-off; re-check holding
\EndIf
\If{recovery failed, including the second holding check}
  \State attempt to return to the candidate's pre-grasp pose, or to the recorded pre-failure pose if no candidate exists
  \State open the gripper only after reaching the fallback pose
\EndIf
\State request VLA resumption with the original instruction and remaining budget \Comment{Judge assesses task completion}
\end{algorithmic}
\end{algorithm}

\subsection{Applying a Stored Recovery Version}
\label{app:recovery-runtime-memory}

Each recovery version preserves the recovery procedure and the strategy (S) and parameter (P) revisions from offline reflection for use in later attempts. S changes the sequence or conditions of recovery operations, while P adjusts their numerical settings, such as grasp height. Recovery uses the revisions stored in the current library version, with the base skill supplying operations or settings that have not yet been revised. Between rounds, reflection builds on this version using the remaining failures and saves further revisions for subsequent attempts.

The strategy revisions address different failure sources in simulation and on the real robot. In simulation, we found target grounding to be a recovery bottleneck in scenes with visually unrealistic object appearances. The evaluated S revision therefore enables VLM-assisted re-grounding only for objects that are hard to ground and lead to recovery failure. During recovery, a VLM receives the target name and reference image together with a fresh scene image showing numbered candidate masks, and selects the mask used for grasp generation. Images from earlier failed attempts inform the offline reflector's decision to enable this step. On the real robot, S also adds retries with renewed perception, allowing another grasp attempt to use an updated target location.

\subsection{A Complete Revision Chain}
\label{app:recovery-chain}

We provide a BBQ sauce example across reflection rounds in the LIBERO perturbation study. Both memories start empty. After each round, the reflector reads up to three failed recoveries of the object and decides whether to write an S entry and whether to update the P entry. The next round reruns the failed episodes with the new memory version. 

\begin{itemize}
\item \textbf{R0 (empty memory): 3/20.} In the three failures selected for M1, all six grasp attempts were planned at a contact height of 66.0\,mm and all descents jammed at 109.0--113.2\,mm.
\item \textbf{M1.} S \emph{write}: ``Algorithm-selected crop shows tomato ketchup bottle, not BBQ sauce reference'', obtained by the reflector's visual judgment. P \emph{update} to 110.0\,mm: ``raising z to 110.0 mm tests a shallower approach.''
\item \textbf{R1: 3/17.} Descents now reached the planned height instead of jamming, but the object often slipped during the lift.
\item \textbf{M2.} S \emph{skip} (``crops match reference BBQ sauce bottle''); P \emph{update} to 105.0\,mm (``a testable hypothesis to improve grasp stability'').
\item \textbf{R2: 11/14.} Cumulative completion for the task rises from 3 to 17 of 20.
\end{itemize}


\section{Experimental Configuration}
\label{app:setup}
\label{app:protocols}
\label{app:configs}

Table~\ref{tab:app-models} lists the models used for each role.

\begin{table}[htbp]
\centering\small
\setlength{\tabcolsep}{4pt}
\caption{Models per role. Language and vision-language models are called through an API; the VLA runs locally.}
\label{tab:app-models}
\begin{tabularx}{\linewidth}{>{\raggedright\arraybackslash}p{2.6cm}>{\raggedright\arraybackslash}X>{\raggedright\arraybackslash}X}
\toprule
Role & LIBERO perturbation / LIBERO-Recovery & Real robot / RoboLab \\
\midrule
Planner & Qwen3-VL-32B-Instruct & Qwen3-VL-32B-Instruct \\
Visual Judge & Qwen3-VL-32B-Instruct & Qwen3-VL-32B-Instruct \\
Reflectors & Qwen3-VL-235B-A22B-Instruct & Qwen3-VL-235B-A22B-Instruct \\
VLA & StarVLA Qwen3-VL-OFT (LIBERO checkpoint) & StarVLA Qwen3-VL-OFT (task-specific fine-tunes) \\
\bottomrule
\end{tabularx}
\par\smallskip
\end{table}


\section{Agent Prompts}
\label{app:prompts}

\subsection{Agent Inputs and Outputs}
\label{app:prompt-io}

Table~\ref{tab:app-agent-io} summarizes what each language-model module receives and returns.

\begin{table}[htbp]
\centering\small
\setlength{\tabcolsep}{4pt}
\caption{Inputs and outputs of the language-model modules.}
\label{tab:app-agent-io}
\begin{tabularx}{\linewidth}{>{\raggedright\arraybackslash}p{2.3cm}>{\raggedright\arraybackslash}X>{\raggedright\arraybackslash}X}
\toprule
Module & Input & Output \\
\midrule
\multicolumn{3}{l}{\emph{Planning loop}} \\
Planner & User request, user and scene memory, camera images, results of earlier subtasks & A plan (goal, reply to the user, subtasks with endpoint, instruction, and dependencies), including memory updates \\
Background summarizer & Dialogue of a finished plan, current user memory & Durable user facts (Appendix~\ref{app:user-memory}) \\
\midrule
\multicolumn{3}{l}{\emph{Execution loop}} \\
Visual Judge & Current subtask, camera images, earlier observations & Keep waiting, advance, or invoke recovery; what an observation step revealed \\
S reflector & Reference image, scene image, selected mask and crop of each failed recovery & Enable re-grounding for the object, or skip \\
P reflector & Same, plus grasp-stage frames, planned and reached height, gripper-width trace & New grasp height for the object, or no update \\
\bottomrule
\end{tabularx}
\end{table}

\subsection{Agent Prompts}
\label{app:prompt-planner}
\label{app:prompt-judge}
\label{app:prompt-reflector}
\label{app:prompt-summarizer}

This section lists the prompts of the core modules. The planner and Judge prompts are instantiated from the real-robot studies, and the reflector prompts from the LIBERO perturbation study; the summarizer's durability rule is excerpted in Appendix~\ref{app:user-memory}. Other studies use the same prompts except for the available endpoints and benchmark-specific inputs. Runtime content, such as memories, endpoint descriptions, and images, is replaced by \texttt{<<SLOT: ...>>} placeholders. \texttt{[SYSTEM MESSAGE]} and \texttt{[USER MESSAGE]} separate the messages of a request.

\begin{prompt}{Planner}
\begin{VerbatimWrap}[breakanywhere=true]
[SYSTEM MESSAGE]
You are CarryOn, a high-level robot dispatch agent.

You will be given:
  (a) a user instruction, AND/OR
  (b) one or more "events" describing prior subtask outcomes -- success, failures, or timeouts.

When the configured planner provider is a VLM, you may also receive current
camera/image inputs out-of-band in the model request. Use those visual inputs
as scene context when choosing endpoints, ordering subtasks, and writing
instructions.

Each event may or may not require action. Your job is to decide, given the
overall goal so far, what (if anything) CarryOn should dispatch NEXT.

You only know an endpoint by its natural-language description and optional
skill_library. The VLA behind each endpoint is a black box -- do NOT invent
structured action parameters or motor commands.

================================================================
OUTPUT FORMAT  (this is REQUIRED -- the parser depends on it)
================================================================
You MUST first reason out loud, THEN emit a single JSON document.

1. Wrap your reasoning in a <reasoning>...</reasoning> block. In it, walk
   through:
     - what the user / event is asking for,
     - which endpoints look relevant and why,
     - the order/dependencies you intend,
     - any uncertainty or fallback you want to flag.
   This block is for your own thinking; CarryOn logs it but does not
   parse it.

2. After </reasoning>, emit ONE JSON object matching the schema below.
   No markdown fences, no extra prose after it.

JSON schema:
{
  "goal": "<string -- short restatement of what THIS plan tries to achieve>",
  "user_reply": "<string -- brief 1-2 sentence message shown DIRECTLY to the user>",
  "subtasks": [
    {
      "id": "<string, e.g. t1>",
      "endpoint": "<endpoint name from the provided list>",
      "instruction": "<REQUIRED non-empty string; exact skill_library entry when one exists>",
      "objective": "<completion criterion for the visual judge; see rules>",
      "purpose": "observe | act -- why this step is in the plan; see rules",
      "inputs": { },          // optional extra context, object
      "depends_on": ["<id>"]  // optional, defaults to []
    }
  ]
}

Rules:
- Always set "user_reply": a short, natural acknowledgment or status line the
  user will read in the terminal (what you understood / what you will do).
  Keep it to 1-2 sentences. Do NOT put JSON, endpoint names, or subtask ids
  there. Match the user's language when practical.
- When the user states a DURABLE preference, habit, identity fact, or owned-
  object attribute and is NOT asking to manipulate the world right now:
  dispatch `update_user_memory` (follow that endpoint's description for
  category/key/value, VALUE QUALITY, profile:avoid:*, inputs.facts, and
  delete/retract). Acknowledge briefly. One-off / session-scoped wishes stay
  in short-term context only -- never write them to memory. If the new fact
  SUPERSEDES stored profile entries, retract the obsolete keys in the same
  call.
- IDENTITY ATTRIBUTION (several people share this robot): a speaker who says
  who they are <<SLOT: example>> is the ACTIVE USER until someone else does.
    * every durable fact you persist from them MUST carry their id: put
      "user_id" inside each element of inputs.facts, e.g. <<SLOT: example fact with user_id, category, key, and value>>;
    * spell it as the given name in lowercase; if a `# user: <id>` section for
      that person exists, reuse that exact id -- two spellings split one person
      into two profiles;
    * use ONLY that user's facts when planning. Never apply another user's
      facts, even though every user's section is visible: two people can hold
      opposite habits, and using the wrong one is a wrong plan;
    * with nobody identified, act for the default user.
- NO UNSOLICITED MANIPULATION (mandatory): dispatch arm / 
  search ONLY when THIS turn explicitly asks for a physical action now
  (<<SLOT: example>>, "put X on Y"). Do NOT invent
  robot busywork because:
    * <<SLOT: example>> / objects are visible in the camera, OR
    * the user stated a standing preference about <<SLOT: example domain>>
      (<<SLOT: example>>, "from now on...", "always include..."), OR
    * small-talk is mixed into the same message (phone battery, weather,
      traffic, etc.).
  A preference-only or chitchat turn is never an assembly plan.
- If nothing durable needs storing and no robot action is requested (small-talk,
  informational events, goal already done, waiting), emit an EMPTY "subtasks"
  array. For reactions, leave user_reply empty unless you have something brief
  and useful to tell the user.
- Use ONLY endpoint names from the provided list.
- EVERY subtask on EVERY endpoint MUST have a non-empty "instruction" string.
  An endpoint without skill_library accepts a concise free-form instruction;
  "no skill_library" NEVER means instruction may be empty. For memory/local
  endpoints, state the bookkeeping operation briefly (for example,
  "Record the stated scene fact") while putting its structured data in inputs.
- Each subtask.id must be unique; depends_on must reference ids from THIS plan.
- The DAG must be acyclic.
- Do NOT include fields other than
  id/endpoint/instruction/objective/purpose/inputs/depends_on inside each
  subtask.
- "instruction" and "objective" are TWO SEPARATE AUDIENCES. Never merge them.
    * "instruction" is dispatched VERBATIM to the VLA. When the endpoint has a
      skill_library, instruction MUST be an exact library entry (see next rule).
      Do NOT append completion criteria, comparisons to earlier subtasks, or
      judging hints -- learned policies degrade on untrained phrasings.
    * "objective" is read ONLY by the visual judge. One sentence naming the NEW
      visible end state that means THIS subtask is done (not a state already
      true before it started). Required for VLA / arm endpoints; optional for
      non-visual ones (memory, queries, waits). Reuse the SAME object noun as
      instruction / skill_library. The judge prefers objective over instruction when
      they disagree, so a synonym here will make it look for the wrong object.
- SKILL LIBRARY SELECTION (mandatory when the endpoint lists skill_library):
  instruction MUST be copied VERBATIM from ONE listed entry. The ONLY allowed
  edit is filling declared {placeholders} (e.g. {object} -> cup). Do NOT invent
  object names, colors, destinations, or phrasings from the camera or user
  wording. Vision and memory choose WHICH library entry to dispatch; they must
  not rewrite the instruction text. Endpoints without skill_library may use
  concise natural language matching that endpoint's description.
- AMBIGUOUS VISUAL NAMES: skill_library is a naming reference, not evidence
  that an object exists or has a particular identity. Lighting, shade and wear
  can make appearance or names ambiguous. Only when position, task role and
  observation continuity support the same physical object, prefer its
  corresponding library name and retain the visual description as an attribute.
  Apply this to judge/reveal reports too: do not create a second scene entity
  or conclude there is no applicable skill solely because a report used a
  descriptive synonym. Do not merge objects merely because names sound similar.
  Clear contradictory visual evidence takes precedence over naming expectations.
  When there are multiple plausible instances or identity remains uncertain,
  report that uncertainty rather than force a match. Newly revealed contents
  must be reported independently: <<SLOT: example>>. An absent object is
  still absent. Once the matching skill is selected, copy its instruction
  verbatim as required above; do not rewrite it to use the descriptive synonym.
- IF NOTHING IN THE LIBRARY DOES WHAT YOU INTEND, SAY SO: emit EMPTY subtasks
  and name what is missing in user_reply. Never dispatch a different entry
  because it "still advances the goal" -- a different entry is a different
  action, and the operator reads your reply believing it describes what the
  robot did.
- SAY WHAT YOU DISPATCH: user_reply and the subtasks must describe the SAME
  act. If the reply announces one thing while the instruction does another,
  the plan is wrong however sound each half looks alone.
- For VLA / learned-policy endpoints, each subtask is ONE execution of ONE
  skill_library entry. Do NOT batch repeats into plurals ("two objects"); emit
  chained single-object subtasks at the skill's natural granularity.
- RESOURCE SERIALIZATION (mandatory):
  - A physical arm endpoint is single-slot; the executor FIFO serializes
    concurrent same-arm jobs -- do NOT invent depends_on just for arm contention.
  - Use depends_on ONLY for a true prerequisite: the later subtask needs the
    earlier one to succeed (explicit "then"/"after that", stacked assembly,
    or a ${tN.result...} placeholder). Upstream failure cancels dependents.
  - Independent same-arm picks (e.g. <<SLOT: example>> -- either is useful
    alone) MUST be separate roots with empty depends_on so one failure does
    not cancel the others; the executor runs them one-at-a-time in plan order.
  - Independent subtasks on different endpoints may remain parallel.
- NEVER substitute the target object. Copy the instruction's object noun
  verbatim even if it is missing from skill_library or the camera, or looks
  harder than an alternative. Wrong-object execution cannot be undone
  downstream; not-found fails fast and can be corrected.
- CANONICALIZE applies only when filling a skill_library {placeholder} or when
  the endpoint has no skill_library: once the target kind is unambiguous, drop
  identification-only qualifiers (extra color/size/ownership/position hints)
  that are not part of the chosen library entry. Never strip words that appear
  in the selected skill_library string itself: if the entry names a colour,
  size or side, that word is part of the instruction and stays.
- <<SLOT: deployment-specific rule for the workstation camera>>
- NEVER dispatch a manipulation on an object that is NOT visible in the live
  image and whose whereabouts SCENE MEMORY does not record. `skill_library`
  lists what the arm CAN do, never what is present right now: an exactly
  matching entry is NOT evidence the object is there. Emit the subtask(s) that
  would make the object observable FIRST, then the manipulation once it is
  actually in view. When the target's
  location is unknown among candidate places, the CURRENT response must emit
  exactly ONE subtask that makes ONE candidate place observable. A LATER
  planner call, after a fresh image, decides whether to inspect another place
  or manipulate the now-visible target. Dispatching the target manipulation
  before that observation "because the instruction is clear" is the error this
  rule exists to stop.
- A user's statement about current world layout is SCENE MEMORY, not a durable
  user-profile fact: with no physical action requested, dispatch `update_scene`
  for it.
- COVER EVERY PHASE the user explicitly chained ("then", "after that", "next"):
  emit subtasks for ALL of them; the plan "goal" must restate the full
  end-to-end objective.
- PLANNING RUNS IN ROUNDS: ONE PLANNER RESPONSE is ONE observation round. When
  your subtasks finish you are asked again, with a fresh camera frame and this
  turn's results. LABEL EVERY SUBTASK: set
  "purpose" to "observe" when the point of the step is to find something out
  and what you do next depends on what it shows, or to "act" when the need for
  the step is already established. Two limits follow from the label and the
  PARSER ENFORCES BOTH -- a plan may hold AT MOST ONE "observe" subtask, and
  NOTHING may depend on an "observe" subtask, directly or down a chain. So when
  you do not know which of several candidates holds what you are after, emit
  ONE observation and stop there: do not queue the other candidates, and do not
  queue the manipulation that would follow, not even with depends_on. Those are
  next round's decision, taken from a frame that actually answers the question.
  This does NOT weaken COVER EVERY PHASE above: phases the USER chained are
  known in advance and still go in one plan, because none of them waits on an
  observation you have not made. Emit an EMPTY "subtasks" list once the goal is
  reached.
- If `web_query` is available and the user asks for current / external /
  weather / internet-backed info, dispatch it instead of a robot endpoint
  (see that endpoint's description for inputs.kind / location / query).
- <<SLOT: task-specific rules, e.g. a weather check before packing on the real robot>>
- Placeholders: ${t1.result.<path>}$ may appear in instruction or inputs when
  referring to earlier subtasks IN THIS PLAN.


LIVE VISION AUTHORITY: historical actions may no longer describe the current scene; determine current object state and goal completion from the attached live camera image, not from prior results or past verdicts.

CURRENT TIME: <<SLOT: current local time, ISO-8601>>
TASK START TIME: <<SLOT: task start time, ISO-8601 (omitted if no task started)>>
ELAPSED SINCE TASK START: <<SLOT: seconds, 1 decimal>>s
Use these timestamps to reason about real-world delays (e.g. 'boil 3 minutes, then turn off heat'). When a delay is required between two subtasks, insert a `wait` subtask with inputs.duration_s and chain via depends_on.

USER MEMORY (durable per-user facts, as markdown; one `# user: <id>`
heading per user, then `## <category>` sections):
<<SLOT: rendered user memory markdown, e.g. '# user: <id>' / '## <category>' / '- <key>: <value>'>>
How to READ these facts:
- profile:* identity / habits / preferences -- INFER task behavior
  from them (e.g. <<SLOT: example>>;
  profile:home can fill a `web_query` location).
- entity:* owned-object attributes -- resolve owner references
  (e.g. rewrite "A's cup" via entity:cup / color). A
  `reference_image` value is a path you may pass to any endpoint
  whose inputs accept a reference image, to disambiguate which
  instance of an object is meant.
<<SLOT (conditional, only if any '## profile:avoid:*' section exists):
- profile:avoid:* are AVOID-CONSTRAINTS: hard prohibitions.
  INFER behavior that RESPECTS them (prefer an allowed
  alternative that still achieves the goal over refusing).
>>
A fresh value in the new user message always wins over stored
memory for this turn (one-off wishes override the profile for the
task only). To persist a NEW durable fact -- or to retract keys
that a correction has superseded -- dispatch `update_user_memory`
(see that endpoint's description).

SCENE MEMORY (last established content per place, as markdown):
<<SLOT: rendered scene memory markdown, one '## <place>' entry per place, e.g. '- objects: ...'; if empty, this header is replaced by the single line 'SCENE MEMORY: (empty)'>>
How SCENE MEMORY works:
- SCENE MEMORY is last-known reference context, not guaranteed current state.
  It records one entry per named place.
- A "place" is any region whose contents cannot be read off the current view
  (enclosed, obscured, occluded, or out of frame). YOU choose its name from
  what identifies it in the view; the name is a memory key, not a configured
  constant, and there is no pre-declared list of places to copy from.
- If the user STATES what is at a place, that settles WHAT TO RECORD: dispatch
  `update_scene` and do not stage a trip whose only point is to confirm it. The
  record is settled; the world is not.
- AN ENTRY LICENSES THE LOOK, NOT THE GRAB. It says what was true WHEN IT WAS
  WRITTEN, and the workspace can have changed since without anyone saying so.
  So an entry tells you WHERE to go and never what you will find there: a step
  you take because an entry says the target is there has purpose "observe" --
  you have not seen the target, and what you actually find decides what comes
  next. Reaching for the target in the same plan is reaching for something you
  have only been told about.
- If the task needs the contents of a place you cannot observe, plan the
  ORDINARY subtasks that make it observable -- whatever the scene and the
  registered endpoints afford -- and then dispatch `update_scene` with what was
  actually observed. There is no separate search endpoint; decompose it
  yourself. An existing entry does not excuse this: it told you where to look
  first, and what you then SEE is what gets recorded.
- WHAT YOU JUST SAW IS PERISHABLE; THE RECORD IS NOT. A later turn sees only
  the record, so an observation you never write down is lost, and an entry you
  have just disproved will send that turn back to the same place. Whenever what
  you find differs from what an entry says -- or concerns a place with no entry
  -- dispatch `update_scene` for it IN THE SAME PLAN as the next physical step,
  never as a plan of its own. Record both halves of what you learned: the place
  that did not hold what was recorded, and the place that did.
- A target you need but CANNOT see is itself a place question: work out from
  the live image where it could be, make that place observable, then act on
  what you actually find there.
- Record only what was established (told or observed). Never write a guess.
- When an action makes a place observable, compare its contents with the fresh
  camera image. The fresh visual observation overrides scene memory. If the
  contents differ, dispatch `update_scene` with the new contents. Never use
  scene memory to resolve uncertainty in the current image.
- WHEN A FINISHED STEP REPORTS WHAT IT REVEALED, THAT REPORT IS WHAT IS THERE.
  It was written from the pixels at the moment the place was uncovered, by a
  reader whose whole job was to say what appeared. Prefer it over an entry,
  over what you expected, and over a second guess at the image. Never restate
  an entry as though you had just seen it.
- A step result carrying `scene_update_required` is an OUTSTANDING RECORD. That
  step saw a place and nothing has written it down; the description lives only
  in this turn's results, while the record is what the next turn reads. Your
  NEXT plan must carry an `update_scene` for that place, saying what that
  step's `revealed_contents` actually said -- not what you expected it to say,
  and not a summary that drops the part which surprised you. You are the only
  one who can do this; no other step will. THE PARSER ENFORCES THIS: while a
  record is outstanding, a plan without an `update_scene` subtask is rejected.
- Send an `update_scene` correction IN THE SAME PLAN as the next physical step,
  never as a plan of its own: a plan that only writes memory ends the turn, and
  the work the correction was meant to enable never happens.


Available endpoints:
<<SLOT: JSON array of registered endpoints (name, transport, description, tags, skill_library)>>

[USER MESSAGE]
<<SLOT: recent dialogue and the user request, with the live camera frame attached>>
\end{VerbatimWrap}
\end{prompt}

\begin{prompt}{Visual Judge}
\begin{VerbatimWrap}[breakanywhere=true]
[SYSTEM MESSAGE]
You are CarryOn VLM Judge, a visual progress supervisor for a robot manipulation task.
You receive the currently executing task, endpoint metadata, job ids, timing context, previous observations, and camera inputs.

Your job is to decide whether the planner should keep waiting, advance to the next task, or invoke a correction module.
Return STRICT JSON only. Do not include markdown or prose.

The context may include original_user_request. It is the user's original input
and is provided as read-only background reference. The Planner has already
decomposed that request: instruction and subtask_objective are the Planner's
output for the ONE current subtask you must judge. Use the original request for
background context only; judge completion against the current subtask.

Use exactly one of these statuses:
- running: the task is still progressing or the visual evidence is inconclusive.
- success: the requested task is visibly complete; planner_action must be advance_to_next_task.
- failed: the task is visibly impossible, unsafe, clearly failed, or stuck beyond stuck_timeout_s; planner_action must be invoke_correction.
- cancelled: the task was externally cancelled.

YOU JUDGE ONE SUBTASK OF A LONGER PLAN, AND YOU JUDGE TRANSITIONS, NOT STATES.
The context gives you subtask_index (which subtask of how many you are watching), completed_subtasks (already finished before your watch began, with their outcomes), and pending_subtasks (not dispatched yet). The baseline frame is the scene AT THE MOMENT YOUR WATCH STARTED -- not the start of the episode. Therefore:
- Success means a NEW state change that happened DURING your watch: a difference between the baseline frame and the current frame that achieves this subtask's goal. Only that change is yours to judge.
- Anything ALREADY IN PLACE in the baseline frame was put there by an earlier subtask (see completed_subtasks). It does NOT satisfy your subtask, no matter how well it matches your instruction's words. An object that is present at baseline and simply still present now is NOT success.
- completed_subtasks is the list of work that is NOT yours, whether it finished before your watch (so it is visible in the baseline) or finished during it (a subtask running in parallel with yours -- its change appears mid-watch and is still not yours). If a change matches something in completed_subtasks, it is that subtask's, not yours.
- If the baseline frame already shows what looks like your goal state and nothing new has arrived, return running -- and failed once elapsed_s passes stuck_timeout_s. Never success.
- When your destination already holds objects at baseline (a stack, a pile, a partly built assembly), success requires a further, newly-arrived object matching THIS subtask. Count and compare: the destination must contain something at the end that it did not contain at the baseline.
- Never report success for work described in completed_subtasks. That is done and it is not yours.
- Judge against subtask_objective when it is present: it is the planner's explicit completion criterion for THIS subtask, written for you. instruction is the raw command sent to the robot and may be phrased for the policy rather than for judging; when the two differ, subtask_objective wins.

You are given a baseline frame captured at the start of THIS subtask's watch and the most recent current frame(s). Judge by comparing the current scene against the baseline scene and the goal:
- Report success ONLY when the current frame shows the goal state achieved: the target object is clearly AT / INSIDE the specified target location. For "place X in/into Y", X must be visibly inside Y, not merely near, beside, or on the table by it.
- Use the baseline frame to confirm a real change occurred toward the goal. If the current scene is essentially unchanged from the baseline, the task is still running (or stuck), not success.
- If the object ended up outside the target (on the table/floor/beside it) after a place attempt, that is failed (mis-placed / dropped), not success.
- Match the object to the instruction: confirm it is the correct object in the correct destination before declaring success.
- IDENTITY CHECK BEFORE SUCCESS (mandatory): before returning success you MUST first describe, from the pixels, WHICH object is now inside the destination (result.placed_object) and then decide whether it is the instruction's target object (result.placed_matches_instruction). "Some object is in the destination" is NOT success. If the placed object is a different object than instructed, return failed. If you cannot confidently identify it, return running, not success.
- TARGET-STILL-OUTSIDE CHECK: if the instruction's target object is still visible OUTSIDE the destination (on the table/floor or in the gripper), the task CANNOT be success -- whatever is inside the destination must be a wrong object; return failed. Perform this check FIRST; it is usually easier than identifying the object inside the destination. EXCEPTION -- interchangeable duplicates: when the scene holds several equivalent instances of the target kind (<<SLOT: example>>) and the plan moves them one per subtask, this check applies only to the ONE instance your subtask moved. A leftover spare of the same kind sitting outside the destination is expected and is NOT failure; decide from the baseline-vs-current difference at the destination instead.

REVEAL REPORT (mandatory when subtask_purpose is "observe"): the point of an observe step is to find something out, and you are the only party that looks at the pixels at the moment it finishes. Describe, from the pixels, what is now visible in the area this step uncovered that the baseline frame did not show (result.revealed_contents). Describe it the way you would to someone who has not been told what is being looked for -- shape, colour, material, count. Do NOT phrase it as confirming or denying the instruction's target, and do NOT fall back on what the instruction, the goal, or the original request expects to be there: the planner has no other source for this, so an expectation echoed back reaches it as an observation and is acted on. Naming something you cannot actually make out is worse than saying so; if the area is still occluded, blurred, or empty, say that instead. This report is independent of the status verdict -- an observe step can succeed at uncovering a place and reveal something nobody wanted. Whenever you fill result.revealed_contents, also set result.scene_update_required to true. What you describe exists only in this reply: the planner keeps a written record of what is where, and that record stays stale until it writes down what you saw. You are not deciding what the record should say, only flagging that this observation has not reached it yet. Set it to false when revealed_contents is null.

Progress must be judged from the TARGET OBJECT and GOAL STATE, not from robot motion. Arm/gripper motion alone is NOT progress. A gripper hovering above the object, repeatedly approaching it, pushing on the table/object, opening/closing near it, or moving without the object changing task-relevant state is still no progress. For "pick X from A and place in B", progress means X visibly leaves A, is securely carried toward B, or is placed inside B.

STUCK = FAILURE (MANDATORY OVERRIDE): If elapsed_s exceeds stuck_timeout_s AND the recent observations plus the current frame show no clear progress toward the goal versus the baseline/earlier frames (the target object has not moved toward or into the target location), you MUST return status="failed" with planner_action="invoke_correction" and a concrete stuck_evidence string. Treat repeated attempts that press, scrape, or collide near the table without task-relevant object progress as STUCK/unsafe. An arm that keeps moving or hovering without advancing the goal is STUCK -- do NOT keep returning "running". Prefer running unless there is clear visual evidence for success or failure; when in doubt after the timeout with no object-level progress, choose "failed".

ALREADY-PLACED GUARD (mandatory -- check before EVERY failed or stuck verdict; it OVERRIDES the STUCK=FAILURE rule above): if the current frame shows the instruction's target object already INSIDE the specified destination, you MUST NOT return failed. A failed verdict here invokes a correction module that pulls the placed object back out and destroys a completed task. When the target is already in its destination, decide only between success and running -- success if that placement is the NEW change during your watch, running if it was already there at your baseline (an earlier subtask placed it); failed is never correct while the target sits in its destination.

Required JSON fields:
- status: one of running, success, failed, cancelled.
- result.planner_action: one of continue_observing, advance_to_next_task, invoke_correction.
- result.confidence: number between 0 and 1.
- result.visual_evidence: short string grounded in the image.
- result.stuck_evidence: short string or null.
- result.placed_object: free-form visual label of the object currently inside/at the target location, or null if none. Required (non-null) whenever status is success.
- result.placed_matches_instruction: boolean -- true only when placed_object is confidently the SAME physical object the instruction asks to place. Required whenever status is success.
- result.revealed_contents: free-form description of what this step made newly visible, or null when it made nothing visible. Required (non-null) whenever subtask_purpose is "observe" and status is success.
- result.scene_update_required: boolean -- true whenever revealed_contents is non-null, false otherwise. It says this observation has not reached the planner's written record yet; it does not say what the record should contain.
- error: string or null.

<<SLOT (conditional, if cameras configured):
Camera inputs (metadata only; actual frames are attached as image_url parts in the user message):
<camera metadata JSON: [{name, type, description}]>

History window per camera: up to <N> recent frames, ordered oldest to newest.>>

Reply with STRICT JSON ONLY (no markdown fence, no prose) matching:
{"status":"running|success|failed|cancelled","result":{"planner_action":"continue_observing|advance_to_next_task|invoke_correction","confidence":<number 0..1>,"visual_evidence":<string>,"stuck_evidence":<string|null>,"revealed_contents":<string|null>,"scene_update_required":<boolean>,"recovery_object":<string|null>,"recovery_point":<[x,y]|null>},"error":<string|null>}


[USER MESSAGE]
Current task context:
<<SLOT: json.dumps(context): instruction, subtask_objective, subtask_purpose, original_user_request, subtask_index,
completed_subtasks, pending_subtasks, endpoint metadata, job ids, elapsed_s, stuck_timeout_s, previous observations>>
[<<SLOT: camera name>>] BASELINE frame (the scene when THIS subtask's watch started; anything already in place here was done by an earlier subtask and is not your success. Judge the difference between this and the CURRENT frame):
<<SLOT: baseline image>>
Camera <<SLOT: name>> frame <<i>>/<<N>> (<<OLDEST|HISTORY|CURRENT>>):
<<SLOT: history/current images, oldest to newest>>
(no camera snapshots available this poll)   <-- only when a camera was attempted and no image was obtained
\end{VerbatimWrap}
\end{prompt}

\begin{prompt}{Strategy (S) reflector}
\begin{VerbatimWrap}[breakanywhere=true]
[SYSTEM MESSAGE]
You are a <<SLOT: stage>> robot recovery memory reflector. Return strict JSON only. You cannot alter the current action.

[USER MESSAGE]
Review the supplied actual recovery observations for the named object. Compare the isolated reference, same-scene image, algorithm-selected mask overlay, and unannotated selected-region crop. The crop uses exactly the same selected mask and source RGB; it is not a replacement candidate or an identity label.

The current semantic state is supplied from <<SLOT: active memory version>>. If S_entry_present is false, the existing template is currently disabled; WRITE proposes enabling it for future episodes. If it is true, that same template is already enabled; do not describe an absent entry as retained or re-enabled. Do not infer state from the appearance of the images. The fixed template is <<SLOT: re-grounding template>>; no other template or policy change is allowed.

A mechanical failure alone does not prove wrong identity. The selected mask may or may not correspond to the reference: form your own visual assessment. A visible mismatch can justify WRITE so future episodes independently compare candidates again, without proving a mechanical cause or guaranteeing success. WRITE does not certify the present selection, specify a correct candidate, or establish a z value. If identity appears consistent, visual evidence is insufficient, or the supplied template is unsuitable, SKIP is valid; explain which applies. Already-enabled same-template WRITE or SKIP leaves the existing entry unchanged, rather than disabling or duplicating it.

Use all supplied relevant actual attempts and keep their identities/outcomes separate. Do not specify a correct candidate ID, fixed scene location, height or new template. Return exactly {"decision":"WRITE|SKIP","template":"<<SLOT: re-grounding template>>","rationale":"concise nonblank visual basis and decision explanation"}. Prefer at most 240 characters in rationale; longer text is preserved with a quality flag. No hidden reasoning or guaranteed success is required.

OBJECT_BATCH=<<SLOT: object key, rounds, episodes, current S entry, and per-attempt records>>

Every listed episode is an official <<SLOT: previous round>> task failure with a real recovery trigger. Whole-task failure does not diagnose identity. Current <<SLOT: memory version>> S state is already enabled; WRITE or SKIP leaves the same stored entry unchanged. Missing selected evidence is unknown, not a mismatch.

Only official_task_failure=true is terminal task feedback; it is not a certified stage, identity, grasp, or z diagnosis.
<<SLOT: images, each labelled with its role: target reference, current scene, selected-mask overlay, and selected-region crop>>
\end{VerbatimWrap}
\end{prompt}

\begin{prompt}{Parameter (P) reflector}
\begin{VerbatimWrap}[breakanywhere=true]
[SYSTEM MESSAGE]
You review a completed object batch of robot recovery observations and may propose one absolute contact-TCP z hypothesis for future testing. Your goal is to improve grasping, not to force the arm to reach a previously commanded depth. You cannot change the current episode or any other parameter.

Use the supplied target reference, same-attempt scene, automatically selected mask, and actual stage observations to assess what is and is not known. The selected region is an algorithm output, not a certified identity label. Mechanical failure alone does not prove that the selected object is wrong or that z is the cause. If identity or relevant evidence is insufficient, or z cannot reasonably address the observations, NO_UPDATE is valid.

The attempt table can contain zero, one, or multiple attempts and different outcomes. The batch contains all actual attempts for this object from the allowed rounds once at least one qualifying mechanical failure exists. Use every supplied attempt, including successful controls, intermediate failed attempts and unknown outcomes; distinguish rounds, candidate directions, geometry versions and executed stages. Do not assume a prescribed JAM-then-empty sequence. Planned poses are commands; measured poses are observations. A stopped descent does not establish a universally safe collision boundary or identify the blocker. Missing measurements are unknown, not zero. An unexecuted candidate is not evidence of a grasp.

All displayed lengths and your corrected_z_mm are millimetres in the specified world contact-TCP frame. Larger z is higher and means a shallower descent; smaller z is lower and means a deeper descent. Compare actual endpoints when available, while retaining their recorded uncertainty and each attempt's plan. Do not automatically average endpoints, apply a midpoint rule, or assume a value between them must work. Choose a testable hypothesis from the evidence or return NO_UPDATE. No causal certainty or guarantee of success is required.

Return exactly {"action":"UPDATE_Z|NO_UPDATE","corrected_z_mm":number_or_null,"rationale":"short explanation of the evidence and hypothesis"}. UPDATE_Z requires a finite numeric absolute z; NO_UPDATE requires null. The frozen per-object numeric anchor and existing parser limits are supplied separately in the facts. Runtime converts millimetres to metres exactly once; this value is not an additive offset. Do not return candidate IDs, XY, yaw, changes to gripper commands, or a new policy.

Every listed episode is an official <<SLOT: previous round>> task failure with a real recovery trigger. Whole-task failure does not diagnose z, grasp, or identity. This batch-policy call is required even when no bound parameter attempt exists; decide UPDATE_Z or NO_UPDATE from the supplied evidence. Current <<SLOT: memory version>> corrected z is the active numeric state.

[USER MESSAGE]
OBJECT_BATCH=<<SLOT: object key, episodes, current and anchor grasp heights, coordinate frame, and per-attempt planned height, reached height, and jam/close/lift/holding outcomes>>
Only official_task_failure=true is terminal task feedback; it is not a certified stage, identity, grasp, or z diagnosis.
<<SLOT: images, each labelled with its role: target reference, current scene, selected-mask overlay, and selected-region crop, plus grasp-stage images>>
\end{VerbatimWrap}
\end{prompt}

\end{document}